\documentclass{article}
\usepackage[utf8]{inputenc}
\usepackage{amsthm}
\usepackage{amsmath}
\usepackage{amssymb}
\usepackage{natbib}
\usepackage{mathrsfs}
\usepackage{thmtools}
\usepackage{nameref,hyperref,cleveref}
\usepackage{graphicx}
\usepackage{bbm}
\usepackage{relsize}
\usepackage{floatrow}
\usepackage[label font=bf,labelformat=simple]{subfig}

\usepackage{tikz}
\usepackage[margin=1in]{geometry}
\usepackage{setspace}
\setcitestyle{authoryear, aysep={,}, open={(},close={)}}
\usetikzlibrary{quotes,angles}

\newcommand{\la}{\left \langle}
\newcommand{\ra}{\right \rangle}

\newcommand{\spa}{\textup{span}}

\newcommand{\ones}{\mathbf{1}}

\newcommand{\abs}[1]{\left|#1\right|}
\newcommand{\norm}[1]{\left\Vert#1\right\Vert}

\newcommand{\bigO}{\mathcal{O}}

\newcommand{\1}{\mathbbm{1}}

\newcommand{\mhat}{\hat\mu}

\newcommand{\hcR}{\widehat{\mathcal{R}}}

\declaretheorem[numberwithin=section]{theorem}
\declaretheorem[numberwithin=section,name=Assumption,refname={Assumption},Refname={Assumption}]{assumption}

\declaretheorem[numberwithin=section,name=Lemma,refname={Lemma},Refname={Lemma}]{lemma}

\declaretheorem[numberwithin=section,style=definition,name=Definition,refname={Definition},Refname={Definition}]{definition}

\begin{document}
\title{Implicitly Maximizing Margins with the Hinge Loss}
\author{%
  Justin Lizama \\
  \texttt{jlizama2@illinois.edu} \\
  University of Illinois at Urbana-Champaign \\
}
\date{}

\newbox{\bigpicturebox}

\maketitle

\begin{abstract}
A new loss function is proposed for neural networks on classification tasks which extends the hinge loss by assigning gradients to its critical points.  We will show that for a linear classifier on linearly separable data with fixed step size, the margin of this modified hinge loss converges to the $\ell_2$ max-margin at the rate of $\mathcal{O}( 1/t )$. This rate is fast when compared with the $\mathcal{O}(1/\log t)$ rate of exponential losses such as the logistic loss.  Furthermore, empirical results suggest that this increased convergence speed carries over to ReLU networks.
\end{abstract}

\section{Introduction}
The vanilla empirical hinge loss risk for a linear classifier is given by
\begin{align}
\hcR_{\textup{hinge}}(w;\beta) = \frac{1}{n} \sum_{i=1}^n \max \{ \beta - \la w ,x_i y_i \ra, 0 \}.
\end{align}
It is widely known that the hinge loss together with the explicit regularizer term $\lambda \norm{w}^2$ maximizes the margin of a linear classifier.  Here however, we study the \textit{implicit} regularization induced by the hinge loss without the addition of this regularization objective.  We do this by introducing a modified hinge loss that we will call the \textit{complete hinge loss}, which converges in margin to the $\ell_2$ max-margin separator at a rate of $\bigO(n/t)$ where $n$ is the number of training examples and $t$ is the number of gradient descent iterations.  This rate is exceptionally fast when compared with exponential loss functions such as the logistic loss whose convergence in margin to the max-margin separator is given by $\bigO( 1/\log t)$ \citep{soudry1710} and $\bigO( \log t / \sqrt{t})$ with normalized gradients \citep{nacson1803}.

Recently efforts have been made to explain the ability of neural nets to generalize well by studying the implicit regularization induced by gradient descent.  \cite{soudry1710,ji1803, ji1810,ji1906,nacson1803} provide convergence analyses for linear models on exponential losses that proves they converge in direction to the max-margin with respect to the $\ell_2$ norm.  These results were further extended by \cite{lyu1906, chizat2002} who provided extensions of the analysis to two-layer ReLU networks, showing that these results do carry over in some form to nonlinear neural networks.

In this direction, this paper seeks to study the convergence properties of linear classifiers independently of the exponential losses.  While it is true that exponential losses play a role in the success of neural networks on classification tasks, neural networks still continue to perform well even when trained with the vanilla hinge loss (\cite{janocha1702}).  For this reason, we propose the complete hinge loss, which completes the hinge loss in the sense that it assigns gradients to critical points of the hinge loss in such a way that, for linear models, convergence to the max-margin separator is completed without the use of any explicit regularizer.

We organize the paper as follows.

In \textbf{section \ref{paramconv}} we give an outline of the proof for the parameter convergence rate of $\bigO(n/t)$ for linear models on the complete hinge loss.  We also include empirical results showing this result holds in practice at the end of the section.

In \textbf{section \ref{neural}} we provide a modified version of the complete hinge loss which can be used with neural networks on data sets which are difficult to separate.  We give empirical results that show the complete hinge loss obtains testing set accuracies that are higher than cross entropy with and without normalized gradients on both CIFAR-10 and MNIST.  

In \textbf{section \ref{discussion}} we close the paper with a discussion of both the theoretical and empirical results, and give some potential directions for future work.

\subsection{Notation and Definitions}

\begin{figure}
\center
\includegraphics[scale=0.25]{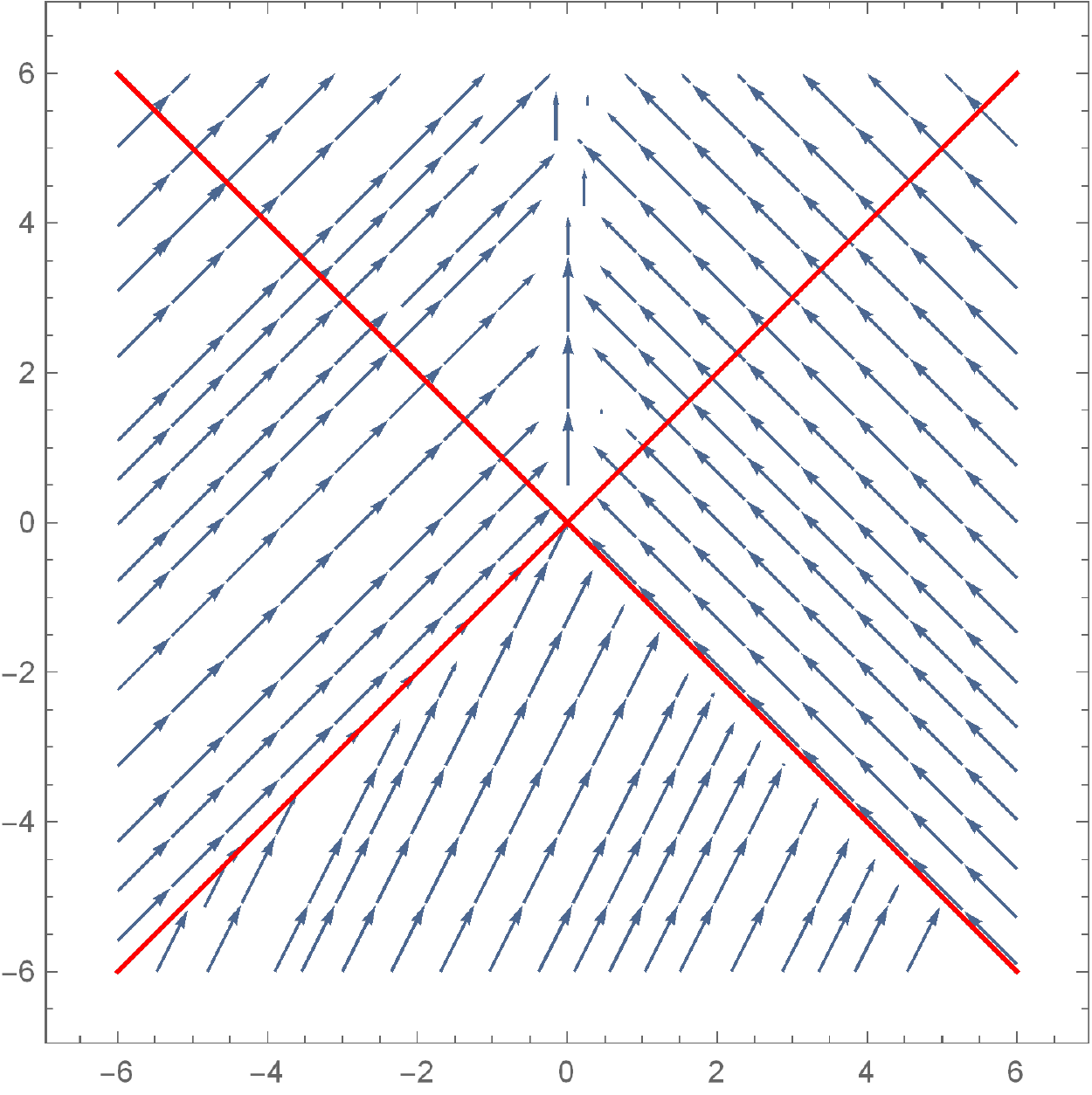}
\includegraphics[scale=0.25]{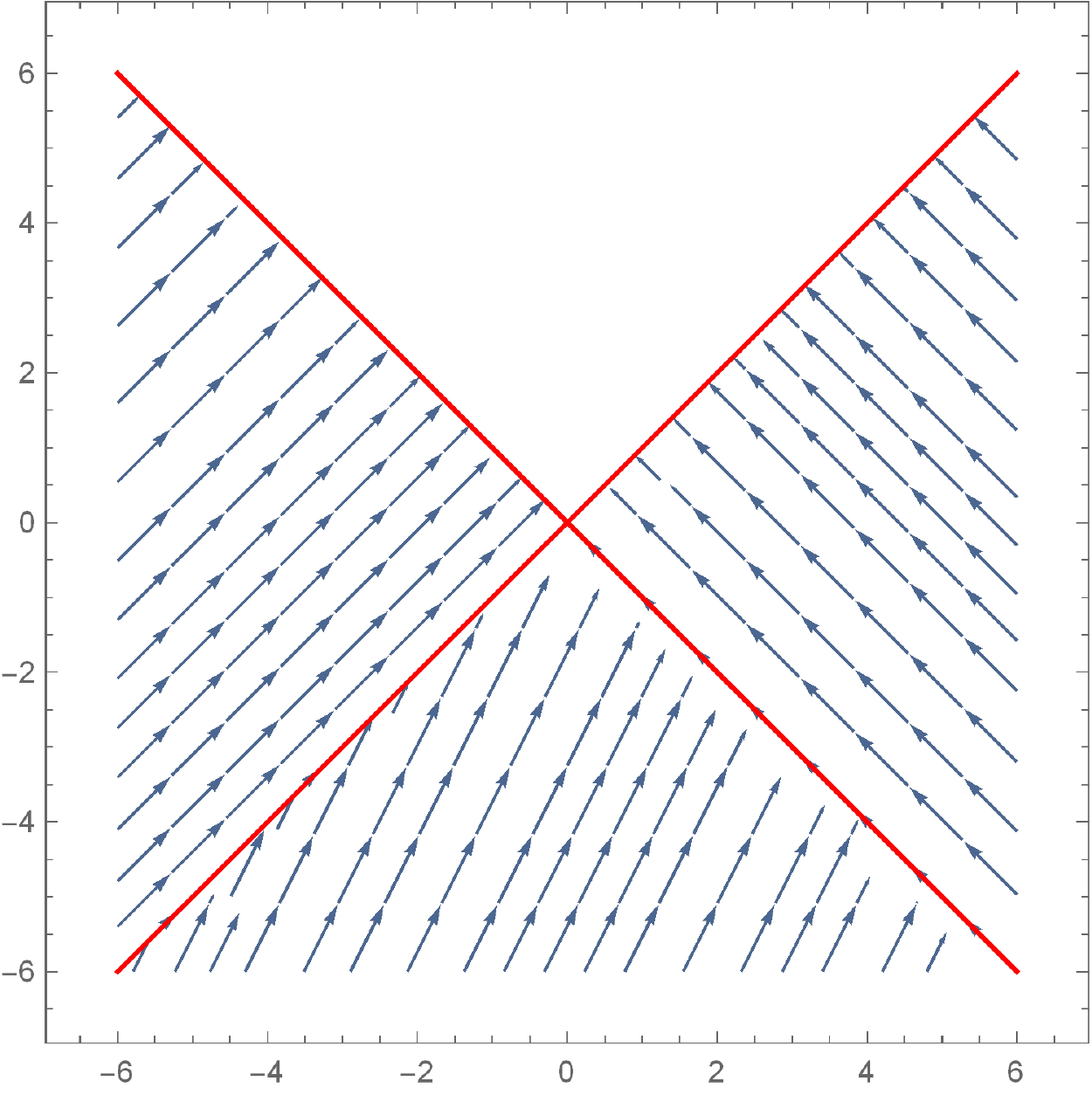}
\caption{To the left we have the gradient flow of the complete hinge $\hcR(f,\beta)$, and to the right we have the gradient flow $\hcR_{\textup{hinge}} (f, 0)$.  The data set used in the figure is $Z = \{(-1,1),(1,1),(2,2)\}$ which has a max-margin separator of $(0,1)$.  As we can see the complete hinge loss is equivalent to the hinge loss except it fills in the blanks and sends the weight vector to the max-margin separator.}
\label{fig:flow}
\end{figure}

We will consider a data set $(x_i,y_i)_{i=1}^n \subset \mathbb{R}^d \times \{-1,+1\}$ with $x_i \in \mathbb{R}^d$ and  $y_i \in \{-1,+1\}$ where $x_i \neq x_j$ for all $i,j$. We will also refer to the set $Z = (x_iy_i)_{i=1}^n = (z_i)_{i=1}^n \subset \mathbb{R}^d$.
\begin{definition}
Define the empirical \textit{complete hinge loss} risk to be the risk function determined by
\begin{align}
\hcR(f,\beta) = -\sum_{i=1}^n \1[f(x_i)y_i \leq \beta]f(x_i)y_i - \1\left[ \sum_{i=1}^n \max \left\{\beta - f(x_i)y_i , 0 \right\} = 0 \right]\frac{\alpha}{\eta} \beta,
\end{align}
where $\beta$ and the parameters of the model $f : \mathbb{R}^d \to \mathbb{R}$ are parameters to be tuned via gradient descent.  $\eta$ is the learning rate, and $\alpha$ is a hyperparameter.  Here we omit the normalization term $\frac{1}{n}$ by absorbing it into the learning rate $\eta$.
\end{definition}
The complete hinge loss can be thought of as a completion of the vanilla hinge loss \begin{align*}
\hcR_{\mathrm{hinge}}(f,\beta) = \sum_{i=1}^n \max \{\beta - f(x_i)y_i, 0\}.
\end{align*}
That is, it assigns gradients to the critical points of $\hcR_{\textup{hinge}}$. For the linear case when $f(x) = \la u ,x \ra $, the gradients are assigned in such a way that convergence to the max-margin is attained. Figure \ref{fig:flow} shows this phenomenon. This complete hinge loss can equivalently be thought of as an infinite sum of hinge loss problems that indefinitely pushes $f(x_i)y_i \to \infty$.  
To better explain how it works, we start out with an initialization $u_0$, then we minimize the hinge loss for $\beta = 0$.  Then once the hinge loss is minimized at $\beta = 0$ we update $\beta$ by some $\alpha \in \mathbb{R}$ and minimize the hinge loss again with $\beta = \alpha$, then minimize again at $2\alpha,3\alpha,\dots, k\alpha,\dots$, and so on.  We repeat this process indefinitely.

We can also eliminate the $\beta$ parameter to obtain a more pure risk function with an infinite series
\begin{align}
\hcR(f) = - \sum_{k=0}^\infty \1 \left[ \min_{(x,y) \in D} f(x)y > (k-1)\alpha \right] \sum_{i=1}^n \1[ (k-1)\alpha \leq  f(x_i)y_i \leq k\alpha ] f(x_i)y_i.
\end{align}

For linear models with $f(x) = \la u ,x \ra $ we will denote the gradient descent iterates at iteration $t$ by the parameter vector $u_t \in \mathbb{R}^d$ and parameter $\beta(t) \in \mathbb{R}$, and we will abuse notation a bit to let $\hcR(u) := \hcR(\la u, \cdot \ra, \beta ) $.  In all experiments and derivations we assume these iterates $(u_t)_{t \geq 0}$ are constructed via vanilla gradient descent  $u_t := u_{t-1} - \eta \nabla \hcR(u_{t-1})$ with a constant learning rate $\eta$.  Additionally, we define a sequence of sets $(S_t)_t$ such that the set $S_t := \{z \in Z ~|~ \la u_t, z \ra \leq \beta(t) \}$ consists of all points trained on at the time step $t$.  That is, we define $S_t$ such that for all $t$, we have $\nabla \hcR(u_t) = \sum_{z \in S_t} z$.

\begin{definition}
Let $\bar u = \arg \max_{\norm{u} = 1} \min_{z \in Z} \la u ,z \ra$ be the $\ell_2$ max-margin separator of $Z$ with margin $\gamma =  \max_{\norm{u} = 1} \min_{z \in Z} \la u ,z \ra = \la \bar u , z \ra > 0 $.  Then the data points $s \in Z$ with $\la \bar u , s \ra = \gamma$ are called the \textit{support vectors}.  We will denote by $S \subset Z \subset \mathbb{R}^d$ the set of support vectors. 
\end{definition} 

\begin{definition}
Let $S$ be the set of support vectors for a linearly separable data set $Z$.  Let $k = \dim \spa (S)$, and let $\Gamma = (\gamma_i)_{i=1}^k \subset S$ be a set of $k$ arbitrarily selected support vectors. We will refer to the matrix $\Gamma = (\gamma_1,\dots,\gamma_k)^\top \in \mathbb{R}^{k \times d}$ as the so-called \textit{support matrix} of the data set $Z$.   We will also use $\Gamma^\dagger = (\gamma_1^*,\dots,\gamma_k^*)$ where $(\gamma_i^*)_{i=1}^k$ are the associated biorthogonal functionals.
\end{definition}
By abuse of notation we will let $\Gamma$ denote both the matrix $\Gamma = (\gamma_1,\dots,\gamma_k)^\top \in \mathbb{R}^{k \times d}$ and the set of row vectors $\Gamma = (\gamma_i)_{i=1}^k$.  Similarly, we will let $\Gamma^\dagger$ denote the set of column vectors $(\gamma_i^*)_{i=1}^k$.  We will also let $\pi_{u}(v) = (I - \hat u \hat u^\top) v$ denote the projection of $v$ onto the orthogonal complement $u^\perp$ where $\hat u = \frac{u}{\norm{u}}$.  The set $H^{\beta}_{u}$ will be the hyperplane  $\{ x \in \mathbb{R}^d ~|~ \la u, x \ra = \beta \}$.  Whenever $\norm{\cdot}$ is used, it refers to the $\ell_2$ norm only.
\section{Parameter Convergence Rate (Linear Models)}\label{paramconv} In this section, we will show that for linear models on linearly separable data the gradient descent iterates of the complete hinge loss converge in direction the max-margin separator $\bar u = \arg \max_{\norm{u} = 1} \min_{z \in Z} \la u ,z \ra$ with margin $\gamma =  \max_{\norm{u} = 1} \min_{z \in Z} \la u ,z \ra = \la \bar u , z \ra > 0 $.   Here we only cover the main results and a proof sketch. We give all deferred proofs for this section in Appendix \ref{appendixA}.

In order to prove the main result we need to first list a couple of assumptions on the data set that make this possible. 
\begin{assumption}\label{supportspan}
If $Z \subset \mathbb{R}^d$ it will be assumed that the support vectors of $Z$ span $\mathbb{R}^d$.
\end{assumption}
\autoref{supportspan} was also applied in prior works \citep{soudry1710, ji1810}, and is true in many cases. This assumption is also justified through application of Lemma \ref{coord}, which allows us to project into the subspace spanned by the support vectors if need be, and Lemma \ref{posgrad}, which allows us to ignore non-support vectors which may lie outside of the subspace.

\begin{figure}
\center
\includegraphics[scale=0.3]{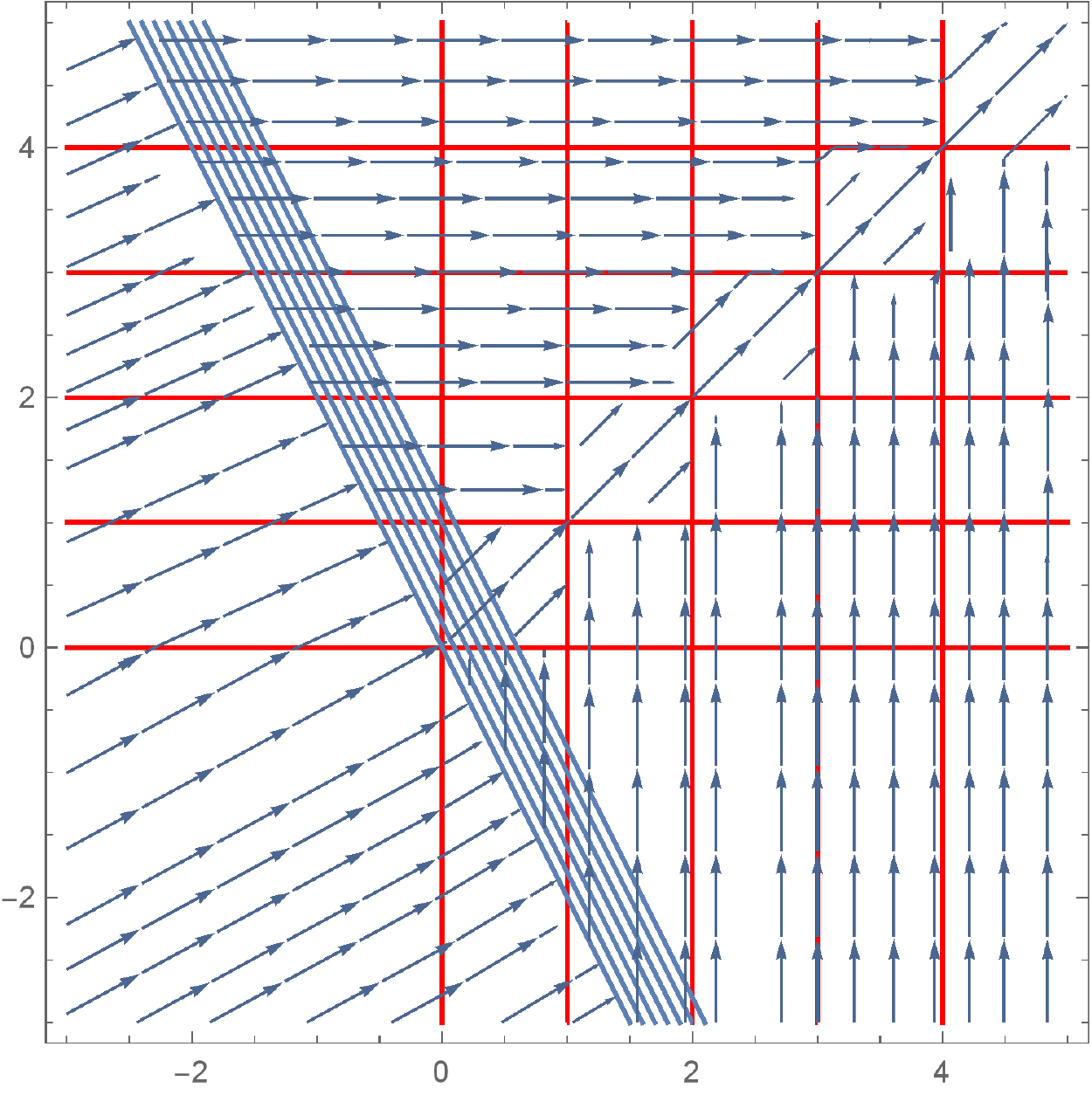}
\includegraphics[scale=0.3]{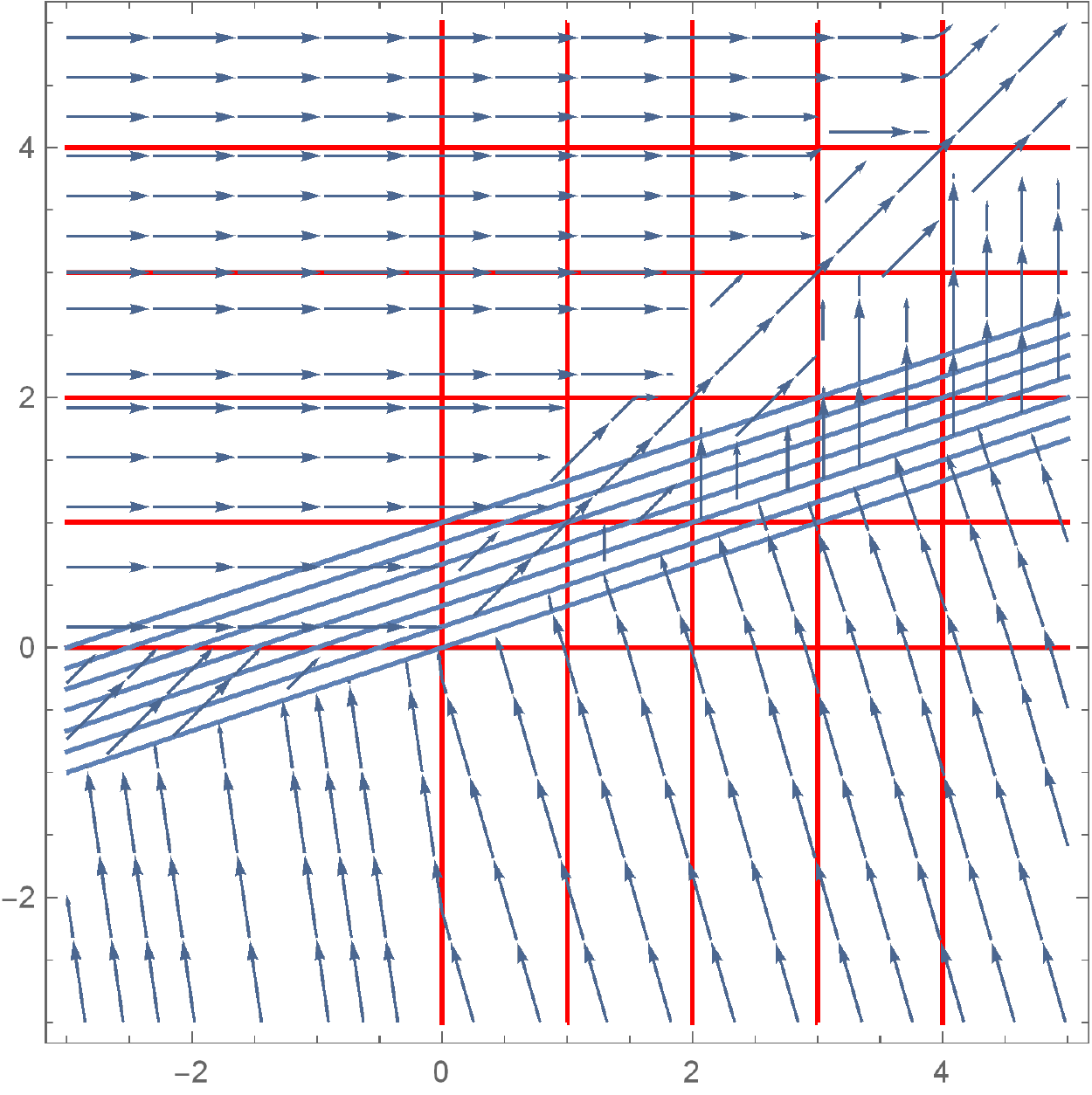}
\caption{On the left is the gradient flow of the complete hinge loss over the data set $Z = \{e_1,e_2, (10,5)\}$ with max-margin $(1,1)/\sqrt{2}$.  The red hyperplanes are $H^{\beta}_{e_1}, H^{\beta}_{e_2}$ and the blue hyperplanes are $H^{\beta}_{(10,5)}$ where $\beta \in \{0,1,2,\dots,6\}$.     As is shown, the blue hyperplanes for $\beta > 0$ never interfere with the gradient updates because the hyperplanes move too slowly in the direction of the flow.  The training example $(10,5)$ only contributes to the gradient when $\beta = 0 $, whereas the support basis vectors $e_1,e_2$ keep up with the flow of the gradients and continue to impact gradient updates indefinitely.  On the right we have a similar effect with $Z=\{e_1,e_2,(-2,6)\}$. }
\label{fig:planevelocity}
\end{figure}

\begin{assumption}\label{reasonablealpha}
The hyperparameter $\alpha$ is chosen such that $\alpha > \gamma \eta $.
\end{assumption}
\autoref{reasonablealpha} is needed to ensure that the learning rate is sufficiently small with respect to $\alpha$.  If this assumption is violated then the iterates will potentially jump around between the cells depicted in Figure \ref{fig:planevelocity}, and this case is not handled in the following analysis.

\subsection{Max-Margin Separator Properties}

In the following \autoref{coord}, we will reformulate the geometrical problem of finding the max-margin as a solution to a linear system.
\begin{lemma}[Support Matrix]\label{coord}
Suppose we have a data set $Z = (z_i)_{i=1}^n$ with margin $\gamma$. If the set of support vectors is $S$ with $ \dim \spa (S) = k$, then there exists a support matrix $\Gamma = (\gamma_i)_{i=1}^k \subseteq S$ with associated biorthogonal functionals $\Gamma^{\dagger} = (\gamma_i^*)_{i=1}^k$ such that the max-margin separator is given by $\bar u = \gamma \Gamma^{\dagger} \ones = \gamma \sum_{i=1}^k \gamma_i^*$.
\end{lemma}
Application of \autoref{coord} allows us to obtain an explicit formula for $\bar u$ in terms of the data set, and in conjunction with \autoref{supportspan}, gives us an invertible linear map $\Gamma^\top$ between a data set $((\Gamma^{-1})^\top z_i)_{i=1}^n$ with the canonical support matrix $I$ and the original data set $(z_i)_{i=1}^n$ with support matrix $\Gamma$.  The definition of the support matrix $\Gamma$ also yields the following useful result that will play an important role in proving convergence to the max-margin. 
\begin{lemma}
\label{dualpos}
Let $ \bar u $ be the max-margin separator of $Z$ with support matrix $\Gamma= (\gamma_i)_{i=1}^d$, then it follows that for all $\gamma_i \in \Gamma$ we have $\la \bar u, \gamma_i^* \ra  \geq 0$.
\end{lemma}
\autoref{dualpos} is an intrinsic property of max-margin separators that must be satisfied.  If it is not true for a given separator $\bar u$, then it follows a separator with a larger margin exists.  This lemma is necessary in order to prove \autoref{posgrad}.

We give a proof for both Lemmas \ref{coord} and \ref{dualpos} in Appendix \ref{appendixA}.
\subsection{Passing Into a Subsequence}

Suppose we have gradient descent iterates $(u_t)_t := (u_t)_{t \geq 0}$ and $t_k = \min \{k > t_{k-1} ~|~ \beta(t_k) > \beta(t_{k-1}) \}$ with $t_0 = 0$, then we can pass into a subsequence $(u_{t_k})_{k \geq 0}$.  This sequence is simply the subset of iterates in which the $\beta$ parameter is actually updated. We know the $\beta$ parameter will always be updated because the data is linearly separable and so we can always attain $0$ risk with the vanilla hinge loss.  We can also truncate this subsequence further at some $t_K$ to remove iterates $(u_{t_k})_{k<K}$ which are problematic for our analysis. It is going to be more convenient to study this subsequence $(u_{t_k})_{k\geq K}$ instead of the actual gradient descent iterates $(u_t)_t$.  However, we must argue that a change of variable from $t$ to $t_k$ is possible, and will not negatively impact the convergence rate.  We formalize this in \autoref{subseq}.  It turns out that given a fixed initialization point $u_0$ and sufficiently large $K$ we obtain that non-support vectors become inconsequential, and so we can ignore most of the data set.  The intuition for why this is true follows from the fact that as $\beta$ grows, the hyperplanes $H^\beta_s$ for support vectors $s \in S$ will move signficantly faster in the direction of the gradient flow than the hyperplanes $ H^\beta_z$ for $z \in Z \setminus S$ will.  Figure \ref{fig:planevelocity} exemplifies this phenomenon. 

\begin{lemma}[Passing Into a Subsequence]\label{subseq}
Consider the subsequence $(u_{t_k})_k$ with $t_k = \min \{k > t_{k-1} ~|~ \beta(t_k) > \beta(t_{k-1}) \}$, then it follows 
\begin{align}
t_k - t_{k-1} =  \bigO(n).
\end{align}
\end{lemma}

We give the proof for \autoref{subseq} in Appendix \ref{appendixA}.  It allows us to pass into the subsequence $(u_k)_k$ at the cost of multiplying our final convergence rate by $n$. 

Next we claim in \autoref{posgrad} that training examples which are not support vectors will only accelerate convergence to the max-margin separator after a certain number of iterations $K$.
\begin{lemma}\label{posgrad}
Suppose $u_k = u_{t_k} \in \mathbb{R}^d$ with $k  > \frac{\gamma + \epsilon}{\epsilon} + C$ and $z \in S_{t_k} \setminus S$ and $\epsilon = \gamma - \min_{z \in Z \setminus S} \la \bar u ,z \ra $, then it follows 
\begin{align}
\la - \pi_{\bar u}(u_k), z \ra \ge C\alpha.
\end{align}
\end{lemma}
\autoref{posgrad} allows us to ignore non-support vectors if we truncate our iterates at $K = \lceil \frac{\gamma + \epsilon}{\epsilon}\rceil$ obtaining a subsequence $(u_{t_k})_{k \geq K}$.  This is because for $k > K$, non-support vectors only accelerate convergence to the max-margin.  To be more precise, it says that non-support vectors $z$ will point towards the ray $a \bar u$ for $a > 0$, and so adding them to the gradient will only push iterates closer in direction to the max-margin separator.

By Lemma \ref{subseq} and Lemma \ref{posgrad} we can instead study the subsequence $(u_{t_k})_{k\geq K}$ instead of the sequence $(u_t)_t$.  By abuse of notation, we will simply denote the iterates $(u_{t_k})_{k\geq K}$ by $(u_k)_k$, and distinguish between $u_k$ and $u_t$ based only on subscript choice.
\subsection{Main Result}
Now that we have argued that a support matrix $\Gamma$ with desired properties exists and showed that we can pass into the subsequence $(u_k)_k$, we are ready to move on to the proof of the main result. 

The proof sketch proceeds as follows.  First we construct a bounded polytope $R_{d}(k)$ for all $k > 0$ such that it surrounds the max-margin separator $k \alpha  \bar u /\gamma$. Then we will show that if $u_k \in R_d(k)$ for some $k$, then the iterates $u_{k+T}$ will be trapped in $R_d(k+T)$ indefinitely for all $T \geq 0$, and hence we obtain convergence to the max-margin separator.  

\begin{figure}
\center
\includegraphics[scale=0.35]{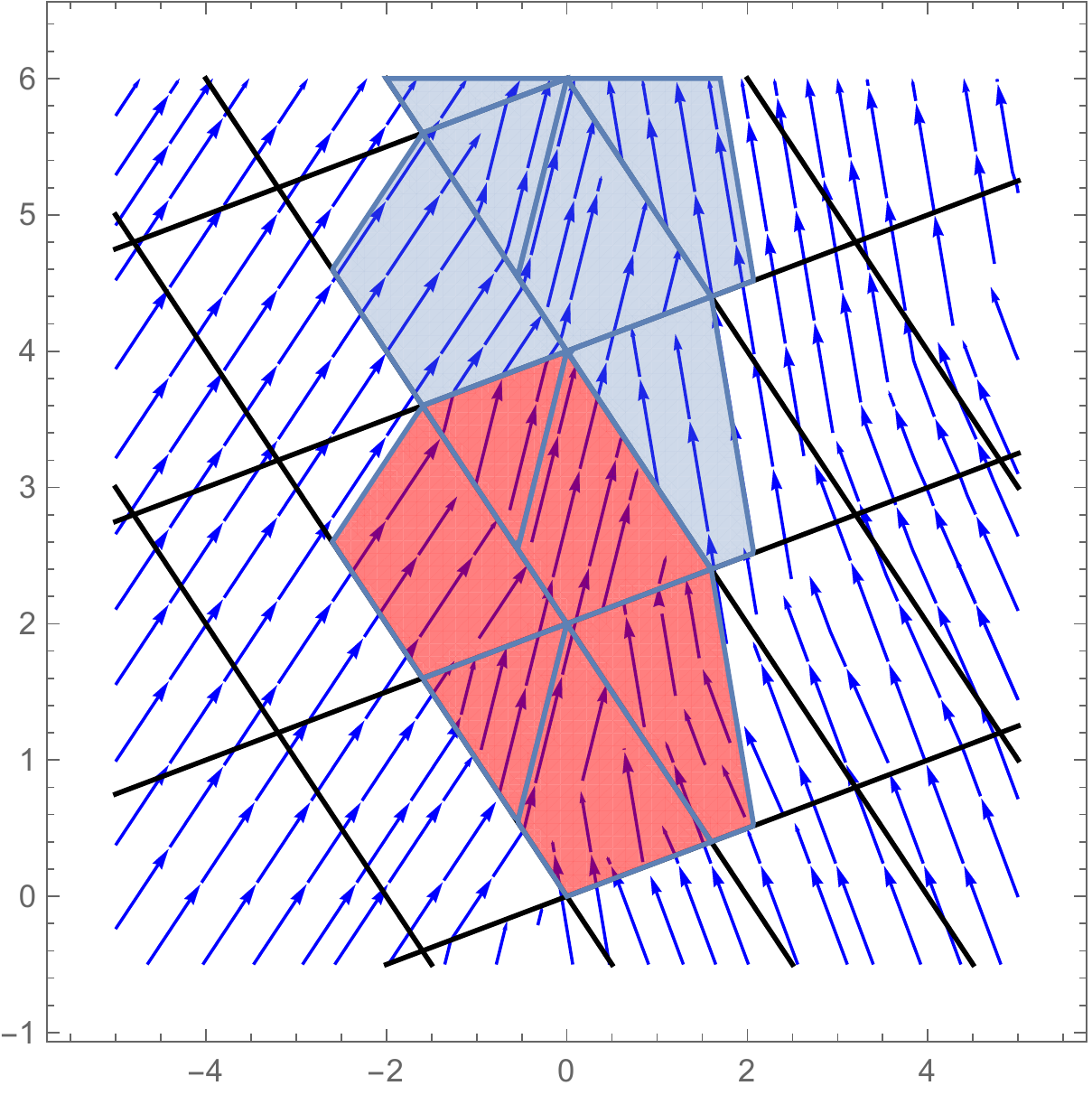}
\caption{The $2$-dimensional overlapping polytopes $R_{2}(1),R_{2}(2),R_{2}(3)$ along with the gradient flow with data set $Z = \{ (1/2,1/2), (-1/8,1/2), (-2,3)\}$ and max-margin separator $\bar u = e_2$. The polytope $R_{2}(1)$ is shaded in red to show the shape of the polytopes.  Also, notice the flow outside of $R_{2}(k)$ points inwards forcing all iterates into $R_{2}(k)$. }
\label{fig:rdk}
\end{figure}

Define the parallelotope $P(k) := \{ u \in \mathbb{R}^d ~|~ (k-1)\alpha \leq \la u, \gamma_i \ra \leq k\alpha, 1 \leq i \leq d\}$, then we will define $R_d(k)$ such that it is a bounded superset of all possible gradient trajectories from any $u \in P(k)$ to $P(k+1)$. A more complicated, but more convenient construction for this polytope is explained in detail in Appendix \ref{appendixA}.  Here in \autoref{polytope} we discuss its properties.
\begin{lemma}\label{polytope}
Under Assumptions \ref{supportspan} and \ref{reasonablealpha}, the bounded polytope $R_d(k) \subset \mathbb{R}^d$ has the following properties:
\begin{enumerate}
\item $\frac{k \alpha}{\gamma} \bar u \in R_d(k)$.
\item If $u_k \in \bigcup_{k \geq 0} R_d(k)$ then $u_{k+T} \in \bigcup_{k \geq 0} R_d(k)$ for all $T \geq 0$.
\item If $u_t \notin R_d(k)$ for some $k$ then there exists a constant $T > 0$ such that $u_{t+T} \in \bigcup_{r \geq 0} R_d(r)$.
\end{enumerate}
\end{lemma}
Property 1 is immediate by definition of $R_d(k)$, but property 2 and 3 require a bit more work.  Figure \ref{fig:rdk} shows the desired polytope $R_d(k)$ for $d=2$ on an example data set. Next, we provide \autoref{stayahead2} which formalizes the phenomenon shown in Figure \ref{fig:planevelocity}.

\begin{lemma}[Support Vectors Stay Ahead.] \label{stayahead2}
Suppose $u_k \in R_{d}(k)$ and let $C = \sup_{x,y \in R_{d}(k)} \norm{x-y}$, then it follows if $k > \frac{\gamma C \norm{z}}{\alpha \epsilon}$ then $\min_{z \in Z \setminus S} \la u_k ,z \ra \geq k\alpha$.  
\end{lemma}

With all supporting lemmas given, we are now ready to state our main result \autoref{linearparamconv}.
\begin{theorem}[Parameter Convergence Rate]\label{linearparamconv}
Suppose $Z$ is linearly separable, $\alpha > \gamma \eta $, and $f(x) = \la u ,x \ra $, then under Assumptions \ref{supportspan} and \ref{reasonablealpha} the gradient descent iterates $u_t$ of the complete hinge loss behave as
\begin{align}\label{margingap}
\gamma - \min_{z \in Z} \la  \frac{u_t}{\norm{u_t}}, z \ra  = \bigO \left( \frac{n}{t} \right),
\end{align}
and consequently,
\begin{align}
1 - \la \frac{u_t}{\norm{u_t}} , \bar u \ra  &= \bigO \left( \frac{n}{t} \right) \\
\norm{ \frac{u_t}{\norm{u_t}} - \bar u } &= \bigO \left( \sqrt{\frac{n}{t}} \right) .
\end{align}
\end{theorem}
The proof of \autoref{linearparamconv} mainly relies on \autoref{polytope}.  By property 2 of \autoref{polytope} we know that after a constant time we will always have $u_k \in R_d(k)$. Also, by property 1, we have $ k\alpha  \bar u /\gamma \in R_d(k)$, and so it follows the distance between $k \alpha \bar u / \gamma$ and $u_k$ is bounded indefinitely.  Then we can apply \autoref{subseq} to switch from the subsequence $(u_k)_k$ back to the original gradient descent iterates $(u_t)_t$ at the cost of multiplying the final rate by $n$.
\subsection{Experiments (Linear Model)}
\begin{figure}[H]
\centering
    \renewcommand\thesubfigure{(\alph{subfigure})}
  
\sidesubfloat[]{\includegraphics[scale=0.3]{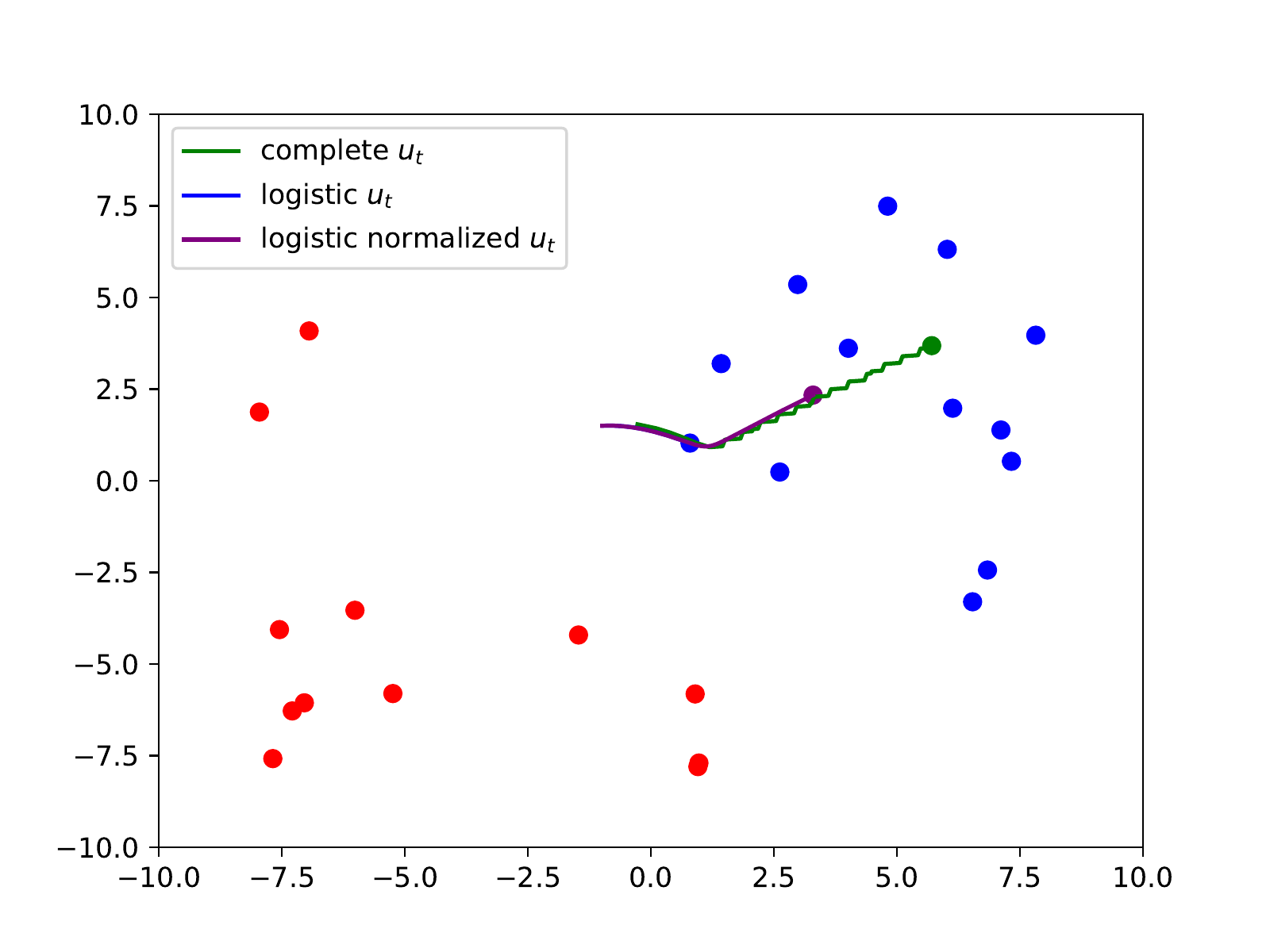}\label{fig:parameterconv:a}}
\hspace*{1em}
\sidesubfloat[]{\includegraphics[scale=0.3]{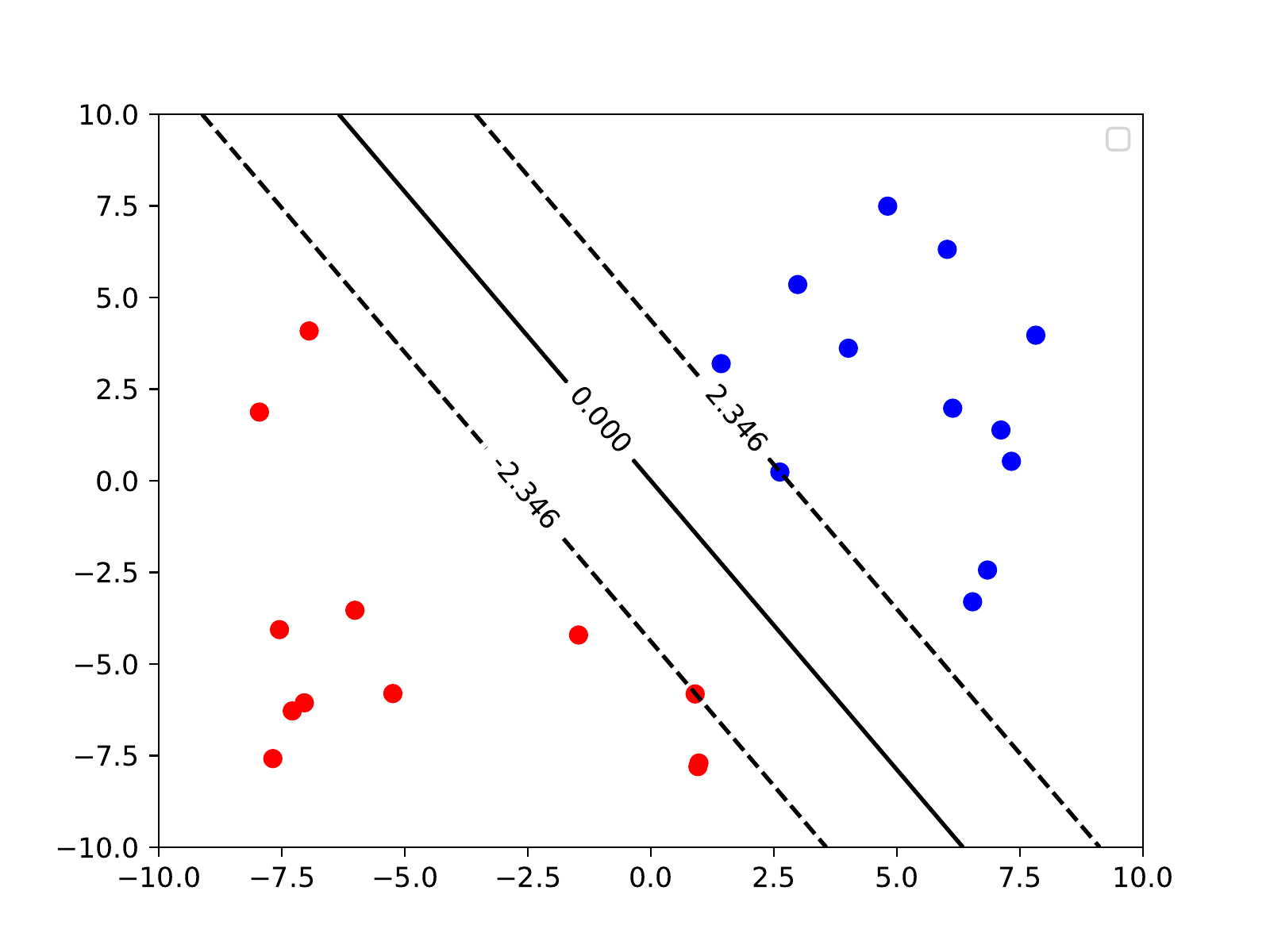}\label{fig:parameterconv:b}}
\caption{Parameter convergence for a synthetic 2-dimensional data set.  Blue and red dots represent training examples with $+1$ and $-1$ labels respectively. \textbf{(a)} The trajectory of the gradient descent iterates $u_t$ for the complete hinge loss,  logistic loss, and logistic loss with normalized gradients all starting at the same initialization point. \textbf{(b)} The separator line obtained by training on the complete hinge loss along with the attained margins.}
\label{fig:parameterconv}
\end{figure}

\begin{figure}[H]
\centering
\renewcommand
\thesubfigure{(\alph{subfigure})}
\begin{tabular}{cc}
\raisebox{40pt}{\sidesubfloat[]{\includegraphics[scale=0.35]{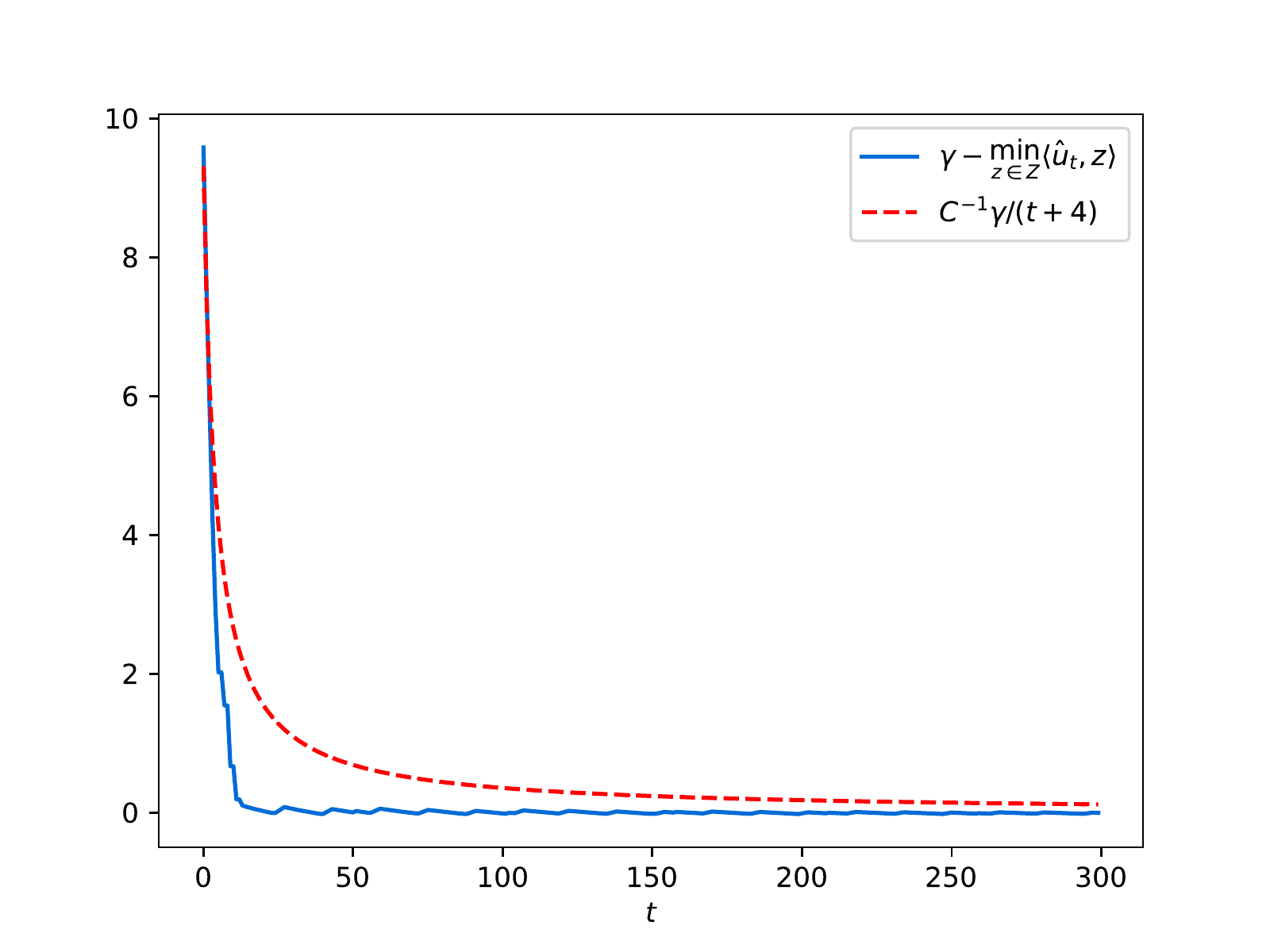}\label{fig:rates:a}}} & \raisebox{-10pt}{
\begin{tabular}{c}
\sidesubfloat[]{\includegraphics[scale=0.2]{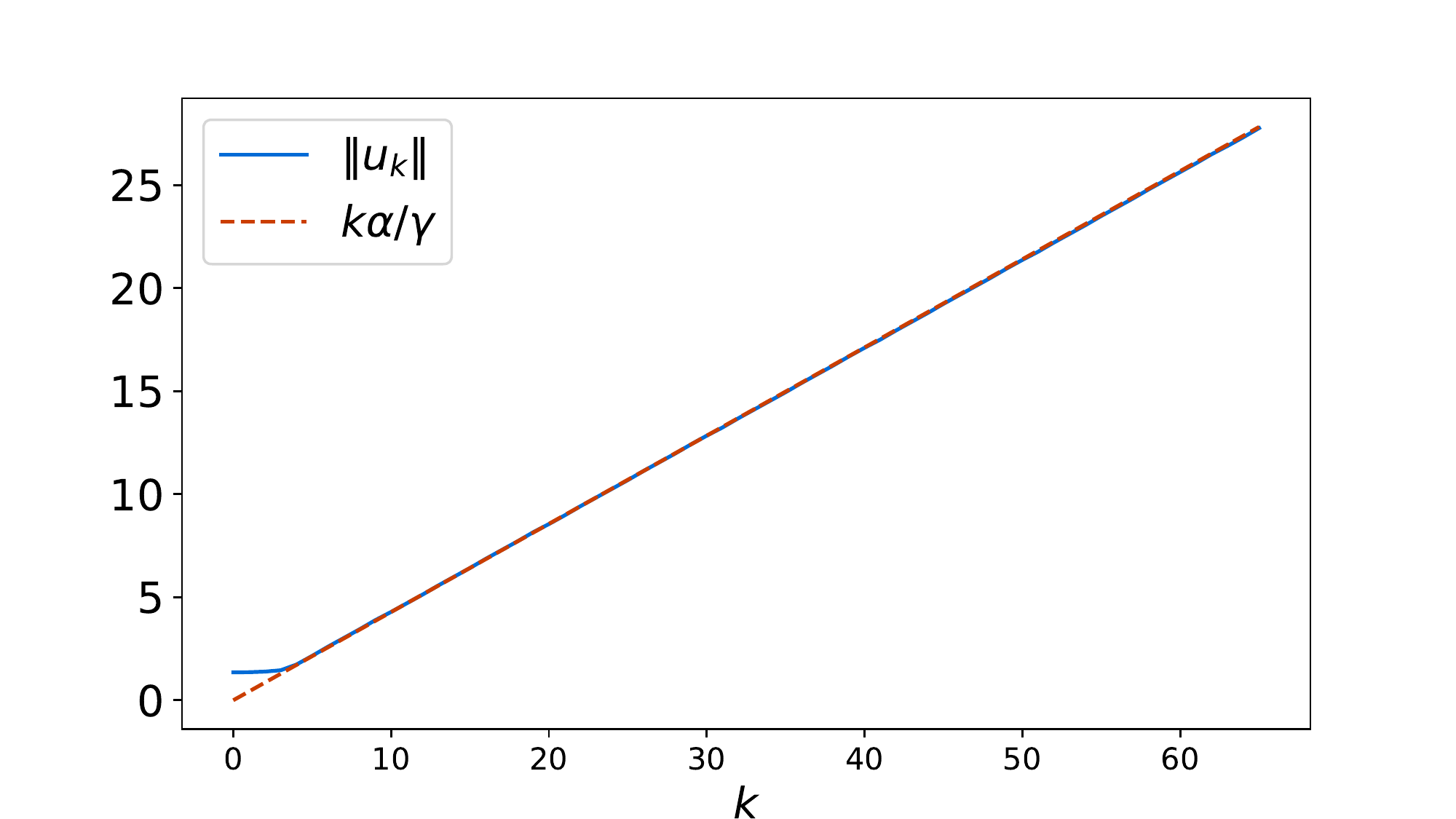}\label{fig:rates:b}}\\
\sidesubfloat[]{
\includegraphics[scale=0.2]{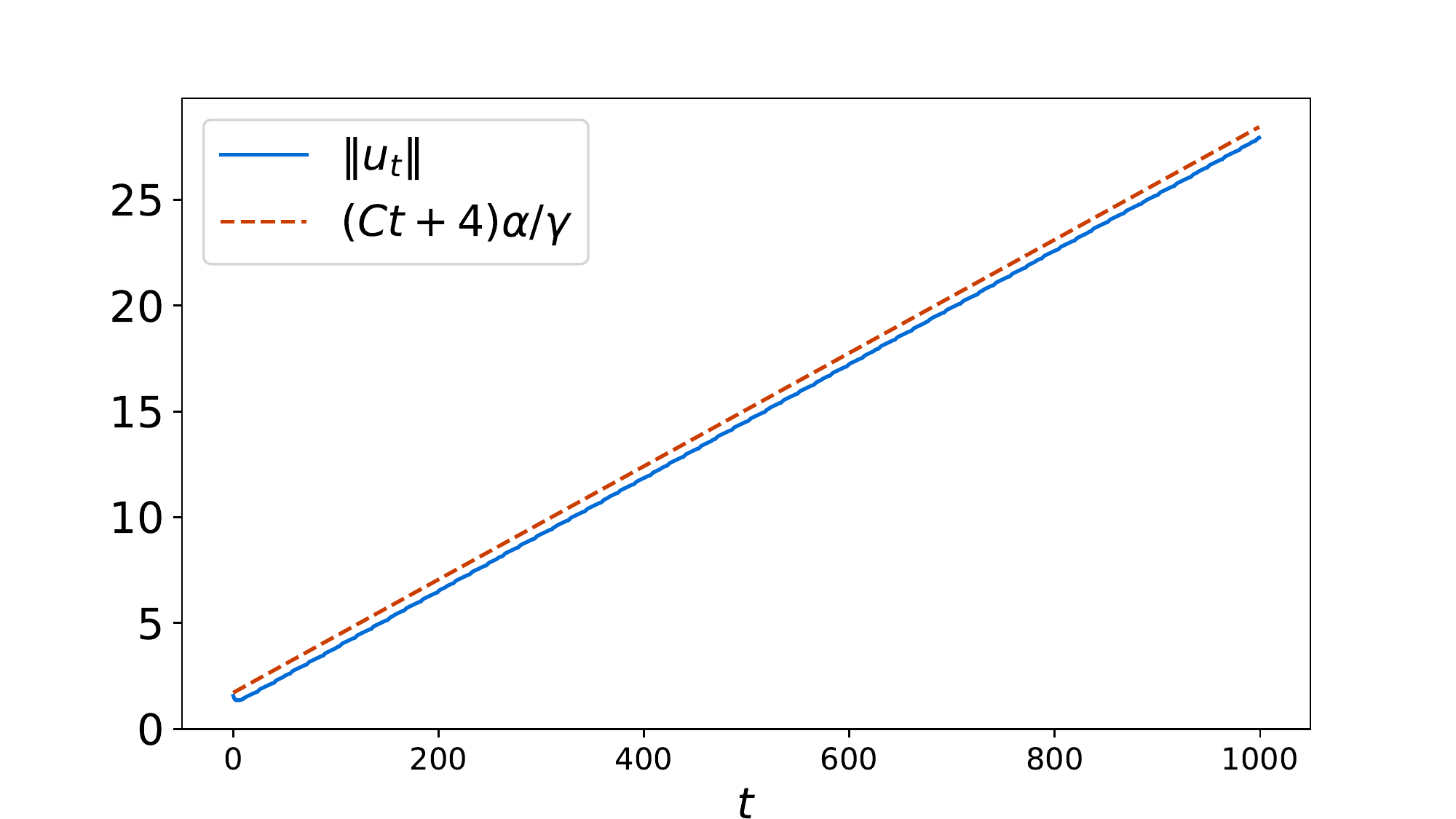}\label{fig:rates:c}}
\end{tabular}}
\end{tabular}
\caption{Convergence rates for linear models on the synthetic data set in Figure \ref{fig:parameterconv}.  The constant $C=  1/16$ (uniquely determined by the data set) is chosen such that the upper bounds hold.  \textbf{(a)} The rate the margin gap (eq. \ref{margingap}) goes to $0$ showing that the margin gap is $\bigO(1/t)$. \textbf{(b)} Plot of $\norm{u_k}$ vs. $k$, showing that $\norm{u_k} = \bigO(\frac{k\alpha}{\gamma})$.  \textbf{(c)} Plot of $\norm{u_t}$ vs. $t$ showing that $\norm{u_t} = \bigO(\frac{t\alpha}{\gamma})$. }
\label{fig:rates}
\end{figure}

Here we give empirical results justifying \autoref{linearparamconv}.  We implement full-batch gradient descent on the synthetic linearly separable data in Figure \ref{fig:parameterconv} using hyperparameters $\alpha = 1$ and $\eta = 0.01$.  In Figure \ref{fig:parameterconv:a} we plot the trajectories of the gradient descent iterates for the complete hinge loss, logistic loss, and logistic loss with normalized gradients.  As we can see the norm of the gradient descent iterates for the complete hinge grow the fastest.  In Figure \ref{fig:parameterconv:b} we show the resultant separator line obtained from optimizing the complete hinge risk $\hcR(u)$.  Figure \ref{fig:rates} shows plots which confirm the margin gap (eq. \ref{margingap}) convergence rate from \autoref{linearparamconv}.

\section{Complete Hinge for Neural Networks}\label{neural}
In this section we discuss empirical results for neural networks on the complete hinge loss. For the experiments we use a modified complete hinge loss with an additional hyperparameter $\zeta$ defined
\begin{align}
\hcR(f,\beta,\zeta) =  - \frac{1}{n}\sum_{i=1}^n \1[f(x_i)y_i \leq \beta]f(x_i)y_i - \1\left[  \sum_{i=1}^n \max \left\{\beta - f(x_i)y_i , 0 \right\} \leq  \zeta \right]\frac{\alpha}{n \eta} \beta.
\end{align}
The purpose of the hyperparameter $\zeta$ is to control when the $\beta$ parameter is allowed to update.  It relaxes the requirement that we need to attain $0$ risk on the hinge loss before we can update the $\beta$ parameter. Also notice that we reintroduce the normalization term $\frac{1}{n}$ here to make scaling consistent with other empirical risk functions used in other literature.  In order to extend the risk function to multiclass classification we use the strategy proposed by \cite{weston1999}.  The details of this extension are given in \autoref{appendixB}.  For data sets that are easy to separate such as MNIST we still use $\zeta = 0$, but for more difficult data sets such as CIFAR-10 a nonzero $\zeta$ will be used.

We compare this modified complete hinge loss  with cross entropy with and without normalized gradients.  We use vanilla gradient descent with a constant learning rate as the optimization algorithm in all experiments.  We run tests on the data sets MNIST (\cite{lecun1998}) and CIFAR-10 (\cite{krizhevsky2009}). 

\begin{figure}[H]
\centering
    \renewcommand\thesubfigure{(\alph{subfigure})}
  
\sidesubfloat[]{\includegraphics[scale=0.3]{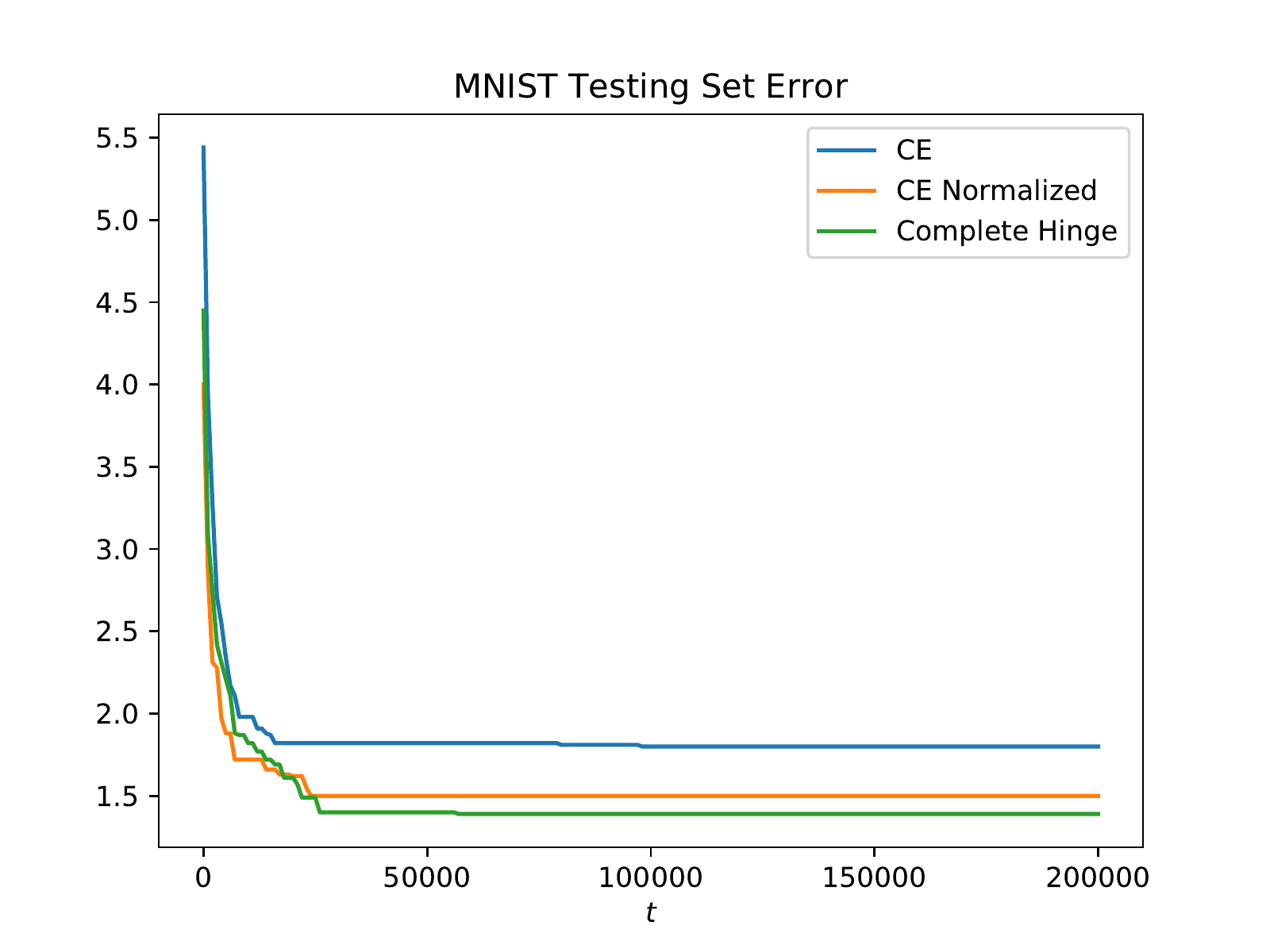}\label{fig:testingerror:a}}
\hspace*{1em}
\sidesubfloat[]{\includegraphics[scale=0.3]{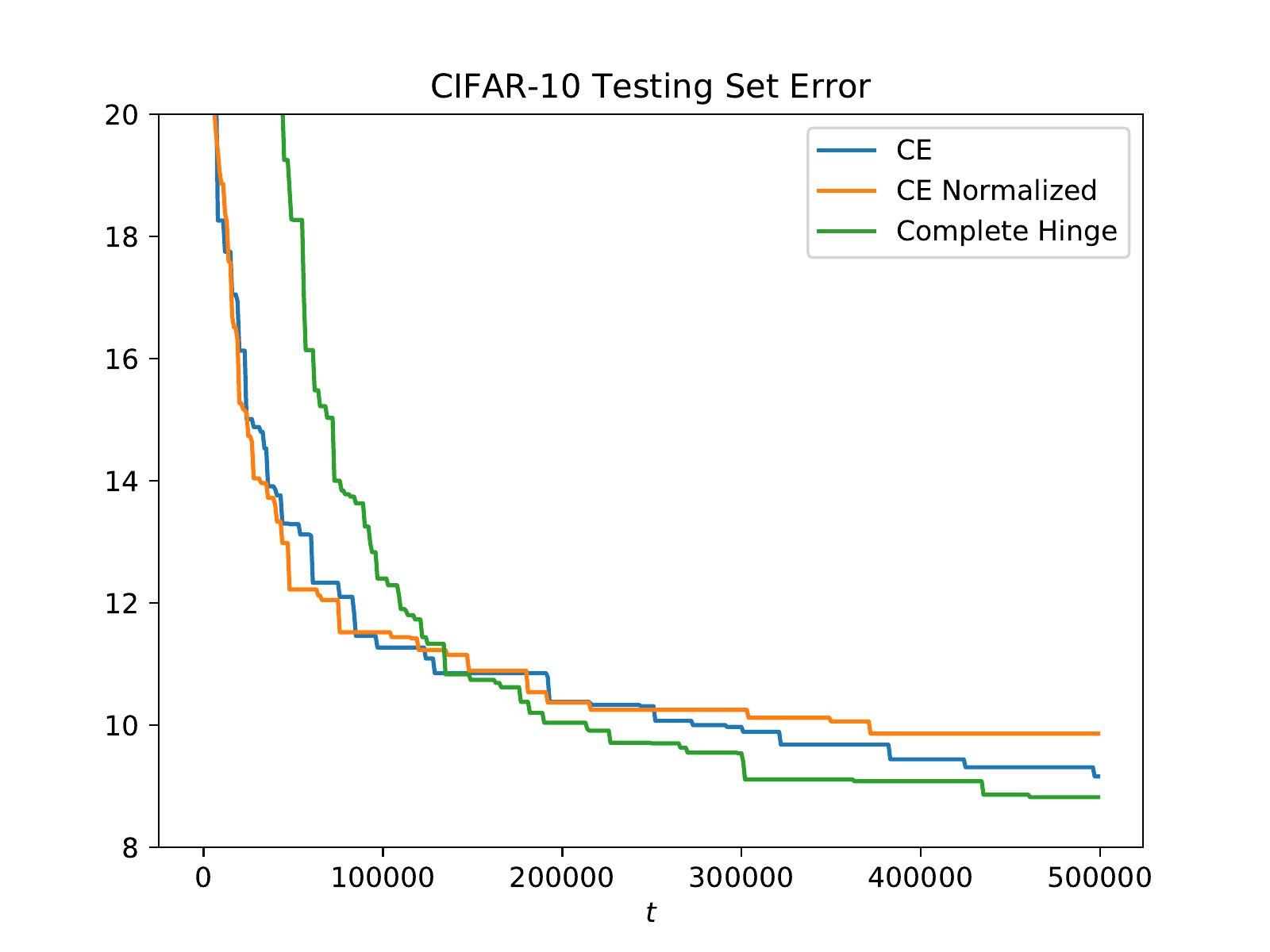}\label{fig:testingerror:b}}
\caption{Plot of lowest yet seen error on testing set over time for \textbf{(a)} MNIST and \textbf{(b)} CIFAR-10.  Cross entropy (CE) with and without normalized gradients is compared with the complete hinge.}
\label{testingerror}
\end{figure}
\subsection{MNIST Experiments}
For MNIST we train a 2-layer neural network with a ReLU activation and 1024 hidden units.  The MNIST data set consists of 70000 grayscale images of digits 0-9 for a total of 10 classes.  We use 5000 images for the validation data, 10000 images for the testing data, and the rest for the training data. 

For both the complete hinge loss and cross entropy we performed a grid-search using the validation data to select hyperparameters $\eta$ and $\alpha$.  We trained for 25000 iterations and compared results on validation data to determine these hyperparameters.  For the complete hinge we chose $\eta \in \{0.01,0.1,0.3,1.0,3.0,5.0\}$ and $\alpha \in \{0.1,1,10,10^2,10^3,10^4\}$, and found $\eta = 0.1$ and $\alpha = 10$ to be optimal.  We used $\zeta = 0$ for this data set.  For cross entropy  we chose from the same range of values for $\eta$ and found $\eta = 0.1$ to be optimal with normalized gradients and $\eta = 1.0$ without normalization.  We use a batch size of 100.

Figure \ref{fig:testingerror:a} shows the ``lowest yet seen'' testing set error over time (i.e. the lowest testing error observed until that point in time).  The model was trained for $2 \times 10^5$ iterations total. The complete hinge loss is able to achieve a testing set classification error of $1.39\%$ whereas cross entropy achieves an error of $1.50\%$ with gradient normalization and $1.80\%$ without gradient normalization.  Thus, the complete hinge achieves a small improvement over cross entropy with normalized gradients. 
\subsection{CIFAR-10 Experiments}
The CIFAR-10 data set consists of 60000 colored images with 10 different classes.  We use 5000 images for the validation data, 10000 images for the testing data, and we use the rest for the training data.  For CIFAR-10 we train a 13-layer convolutional network based loosely on the architecture proposed by \cite{hasanpour1608}.  We give more details on the architecture in Appendix \ref{appendixB}.  

To choose hyperparameters, we trained for 30000 iterations and compared results on validation data.  We did a grid-search for $\eta \in \{0.01, 0.05, 0.1,1.0, 2.0 \}$ and $\alpha \in \{1,10,10^2,10^3\}$. For the complete hinge $\alpha = 10^3$ and $\eta = 0.01$ were best.  However, we used $\alpha = 10$ because the best and second best choices $10^2$ and $10^3$ resulted in exploding gradients without significant gradient clipping.  We chose to decrease $\alpha$ rather than use gradient clipping because later in training gradient clipping resulted in significantly slower convergence.  This is because the gradients grew larger as training went on, and so gradient clipping hampered performance more as time went on.  We used $\zeta = 4$ for both the final test run and during validation evaluation. For both cross entropy methods $\eta = 0.1$ performed best. We used a batch size of 100 and $\ell_2$ weight regularization with a scaling factor of $\lambda = 5 \times 10^{-5}$ for all methods.   

In Figure \ref{fig:testingerror:b} we show the ``lowest yet seen'' error for CIFAR-10 on the testing set for all three loss functions.  We trained for a total of $5 \times 10^5$ iterations for each of the loss functions.  The complete hinge loss performed the best with $8.86\%$ testing error.  Somewhat surprisingly, next best was cross entropy without gradient normalization with $9.16\%$ error, and cross entropy with normalized gradients achieved $9.86\%$ error.  

\section{Discussion}\label{discussion}
In this paper, we have introduced a new loss function, the complete hinge loss, which, for linear classifiers, attains convergence to the max-margin separator at a very fast rate.  Moreover, we have rigorously proved this convergence rate on linearly separable data. Empirical results verify that these convergence properties carry over to neural networks in some form, in the same way the convergence properties of linear models on the logistic loss carry over to neural networks.  Furthermore, based on the experiments on CIFAR-10 and MNIST we see that the complete hinge generalizes better than cross entropy, or is at least able to converge faster than cross entropy.   There are many potential directions for future work.  Here we specify a few.
\\\\
\textbf{Explicit Regularizers.} In this paper we only study the implicit regularization of the complete hinge loss, and explicit regularization was not included in our analysis.  We have used $\ell_2$ weight regularization in our experiments for CIFAR-10, but we did not tune the hyperparameter very thoroughly.    We have also purposefully left out dropout in our experiments since the complete hinge has to be modified in order to accommodate for it. This is because the risk evaluation is determined by the rate the vanilla hinge loss risk goes to 0.   It could be interesting to examine more closely how these explicit regularizers interact with the complete hinge.
\\\\
\textbf{Hyperparameter Choice.} We have run multiple experiments in which we used a grid-search to determine optimal hyperparameters because even with the convergence analysis it is still unclear how best to choose the hyperparameters $\zeta$ and $\alpha$.  Empirical results for CIFAR-10 suggest that a larger $\alpha$ is better for generalization.  We can conjecture that this simply has to do with the growth rate of the $\ell_2$ norm of the iterates, but it is still not fully clear if this is the right explanation.  Also, we were unable to use the larger $\alpha$ values with the complete hinge without experiencing gradient explosion.  Is there some way to circumvent this issue via a larger weight decay, learning rate decay, or some other strategy?
\\\\
\textbf{Momentum.} We did not use momentum based methods for optimizing in any of our experiments.  It might be interesting to see how the addition of momentum affects the analysis and experiments, or if there is some way to modify the complete hinge loss such that the incorporation of momentum leads to improved results.

\bibliographystyle{plainnat}
\bibliography{neurips_complete_hinge}

\newpage
\appendix
\section{Omitted proofs from Section \ref{paramconv}}\label{appendixA}
Before stating proofs we precisely define the sequences $(u_k)_k \subset (u_s)_s \subset (u_t)_t$.
\begin{definition}\label{subseqk}
When we write $(u_t)_t$ we mean the true gradient descent sequence
\begin{align}
u_t &:= u_{t-1} - \eta \nabla \hcR(u_{t-1}) \\
&= u_{t-1} + \eta \sum_{i=1}^n \1[\la u_{t-1},z_i \ra \leq \beta(t-1)] z_i.
\end{align}
\end{definition}
\begin{definition}\label{subseqs}
When we write $(u_s)_s $ we mean the sequence
\begin{align}
u_s := u_{s-1} + \min_{1 \leq i \leq n}\left\{ \frac{k\alpha - \la u_{s-1},z_i \ra }{\la \sum_{z \in S_{s-1}} z, z_i\ra} ~\left|~ \frac{k\alpha - \la u_{s-1},z_i \ra }{\la \sum_{z \in S_{s-1}} z, z_i\ra} > 0 ~ \right\}\right. \sum_{z \in S_{s-1}} z .
\end{align}
We know this is a subsequence of $(u_t)_t$ as a direct result of eq. \ref{proj}. Intuitively, we increment the index $s$ whenever $\nabla \hcR(u_{s+1}) \neq \nabla \hcR(u_{s})$.
\end{definition}
\begin{definition}\label{subseqt}
When we write $(u_k)_k$ we mean the sequence $(u_{t_k})_k$ such that $t_k = \min \{k > t_{k-1} ~|~ \beta(t_k) > \beta(t_{k-1}) \}$.  This is a subsequence of $(u_s)_s$ because the gradient $\nabla \hcR(u_t)$ changes more often than $\beta$ during optimization.  With $k > \frac{\gamma + \epsilon}{\epsilon}$.
\end{definition}

\begin{figure}
\centering
\includegraphics[scale=0.3]{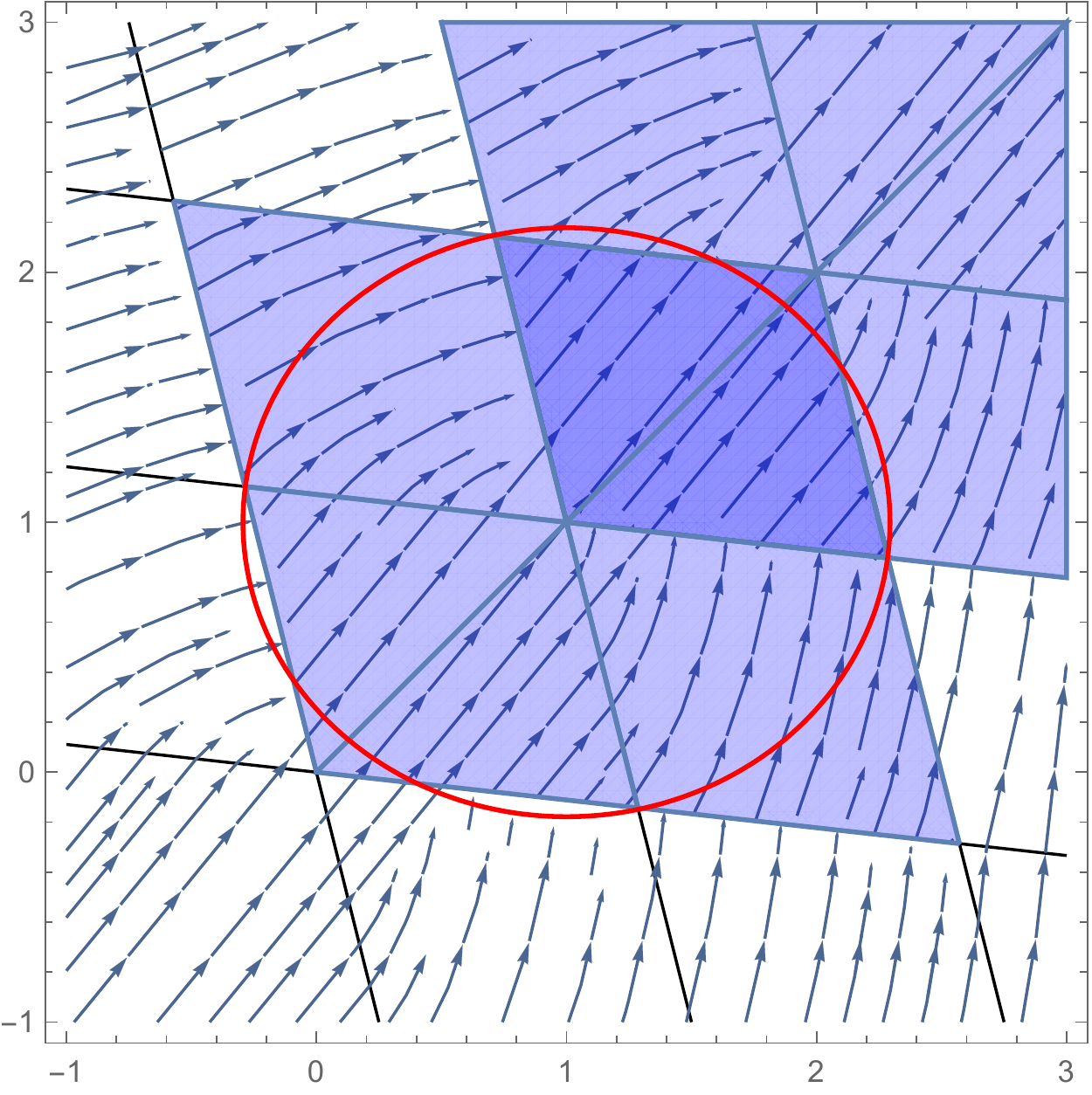}
\includegraphics[scale=0.3]{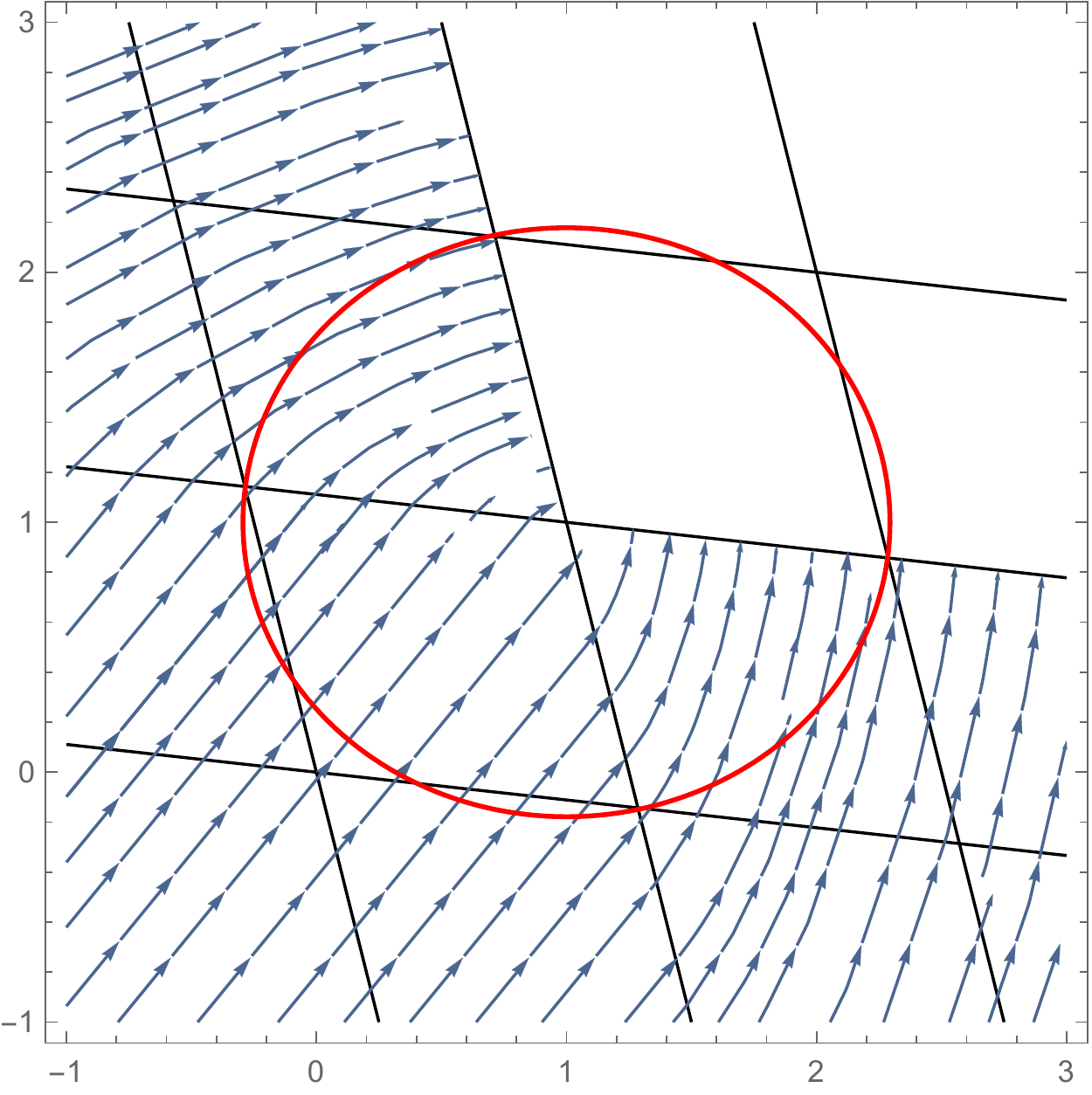}
\caption{On the left, blue shaded region shows a restriction to parallelotope $Q := \{ u \in \mathbb{R}^d ~|~ (k-1)\alpha\ones \preceq \Gamma u \preceq (k+1)\alpha\ones \} $ with smallest $\Gamma$.  Bounding ellipse shows that flow from $P(k)$ to $P(k+1)$ remains contained in $Q$, and hence $R_{d}(k)$ is contained within $Q$. On the right, the vanilla hinge loss for $\beta = 1$, showing the iterates are the same as the complete hinge in $Q$. }\label{fig:reducetosingle}
\end{figure}

\begin{proof}[Proof of \autoref{coord}]
Suppose $\dim \spa (S) = d$, then it follows by \autoref{supportspan} we can choose $\Gamma = (\gamma_i)_{i=1}^d \subset S$ such that the $\gamma_i$ are linearly independent.  Since these are support vectors we must have then that $\Gamma \bar u = \gamma \ones$, and 
\begin{equation}
\bar u = \gamma \Gamma^{-1} \ones = \gamma \sum_{i=1}^d \gamma_i^*.
\end{equation}
Now suppose $\dim \spa (S) = k < d$. Then define $P = (\gamma_1,\dots,\gamma_k)^\top \in \mathbb{R}^{k \times d}$ where $(\gamma_i)_{i=1}^k \subset S$.  Invoking \autoref{supportspan}, additionally assume $P$ is a full rank matrix with linearly independent row vectors.  Then it follows we can transform points $z \in Z$ to construct a new data set $Z_k = (Pz_i)_{i=1}^n$.  In this data set we have support vectors $S_k$ with $\dim \spa (S_k) = k$ and so we can choose $\Gamma_k = P(\gamma_1,\dots,\gamma_k) = P P^\top \in \mathbb{R}^{k\times k}$ and we have as before
\begin{align*}
\bar u_k &=  \gamma \Gamma_k^{-1}  \ones = \gamma (P P^\top)^{-1} \ones .
\end{align*}
The rows of $P$ are linearly independent, so embedding back into $\mathbb{R}^d$ yields
\begin{align*}
\bar u =P^\top \bar u_k = \gamma P^\top (PP^\top)^{-1} \ones =\gamma P^\dagger \ones.
\end{align*}
So then if we take $\Gamma = P$ we obtain the desired result with $\bar u =\gamma \Gamma^\dagger \ones$.
\end{proof}
\begin{proof}[Proof of \autoref{dualpos}]
Suppose $\Gamma = (\gamma_i)_{i=1}^{d-1} \cup \{s\}$ with $\la \bar u, s^* \ra \leq 0$. Then $\bar u =\frac{1}{\norm{\Gamma^\dagger \ones}} \Gamma^{\dagger} \ones $ with $\la \bar u,\gamma_i \ra =\gamma$.  Take the projection of $\bar u$ onto $(s^*)^{\perp}$, $\pi_{s^*}(\bar u)$, then it follows that 
\begin{align*}
\la  \pi_{s^*}(\bar u), \gamma_i \ra &= \la (I - \hat s^*  ({\hat s^*})^\top) \bar u, \gamma_i \ra \\
&=\la \bar u -  \frac{1}{\norm{s^*}^2}\la \bar u,s^* \ra s^*, \gamma_i \ra \\
&= \la \bar u ,\gamma_i \ra  - \frac{1}{\norm{s^*}^2} \la \bar u, s^* \ra \la s^*, \gamma_i \ra \\
&= \gamma.
\end{align*}
And we have
\begin{align*}
\la \pi_{s^*}( \bar u), s \ra  
&= \la \bar u, s \ra - \frac{1}{\norm{s^*}^2} \la \bar u, s^* \ra \la s^*, s \ra \\
&= \gamma - \frac{1}{\norm{s^*}^2} \la \bar u , s^* \ra \\
&\geq \gamma.
\end{align*}
However,
\begin{align*}
\norm{ \pi_{s^*}(\bar u)} \leq \norm{\bar u} = 1 . 
\end{align*}
And so, it follows that the separator $\frac{\pi_{s^*}(\bar u)}{\norm{\pi_{s^*}(\bar u)}}$ has a larger margin than $\bar u$, but this is a contradiction because $\bar u$ is the max-margin separator.  Thus, it follows we must have $\la \bar u ,s^* \ra \geq 0$ and hence if $\bar u$ is the max-margin separator we must have that $ \la \bar u ,\gamma_i^* \ra = \gamma \sum_{j=1}^d \la \gamma_j^*,\gamma_i^* \ra \geq 0$ for all $\gamma_i \in \Gamma$.
\end{proof}
\begin{proof}[Proof of \autoref{subseq}]
By definition $t_k$ is just the points in time when $\beta$ updates.  We know that $\beta(t_k)$ will continue to update indefinitely because the hinge loss $ \ell_{\textup{hinge}} (\la u ,z \ra)=  \max\{\beta - \la u ,z \ra, 0\}$ is convex for any choice of $\beta$, and so gradient descent will find a global minimum every time.  We also know that
\begin{align}
\beta(t_k) = \beta(t_{k-1}) + \alpha = k\alpha.
\end{align}
However, before we can use this change of time variable we wish to upper bound the number of iterations in which each hinge loss optimization subproblem takes to reach a critical point. Let $L$ be the distance traveled from $u_{t_{k-1}}$ to the hyperplane $H_{z}^{k\alpha} = \{x \in \mathbb{R}^d ~|~ k\alpha - \la x,z \ra = 0\}$ at any point in time $t_{k-1}$ and any point $z \in Z$.  Let $\mu := -\nabla \hcR(u_t) = \sum_{i=1}^d \1[ \la u ,z_i \ra \leq k\alpha] z_i$.  Then using a bit of geometry, $L$ can be computed using
\begin{align}
\cos \theta_{\mu,z} = \la \mhat , \hat z \ra = \frac{k\alpha - \la u_{t_{k-1}} , z\ra}{\norm{z} L } .
\end{align}
See Figure \ref{fig:geometry} for a visual representation of this equation. Solving for $L$ yields
\begin{equation}\label{proj}
L = \frac{k\alpha - \la u_{t_{k-1}} ,z \ra}{\la \mhat , z\ra } .
\end{equation}
Gradient descent will go in the constant direction $\mu$ until the distance $L$ is traveled.  This is $S_{t_{k-1}} = \{ z \in Z~|~ \la u,z \ra \leq (t-1)\alpha\}$ is constant for this duration of time.  

Let $(u_s)_s := (u_{t_s})_s$ be a subsequence of $(u_t)_t$ defined such that $u_{t_{s+1}}$ is simply the projection of $u_{t_s}$ onto the nearest hyperplane $H^{k\alpha}_{z}$ for some $z \in Z$ in the direction $\mu$.  We know that the gradient takes step sizes of $\eta \norm{\mu}$, so we have that $t_{s} - t_{s-1}$ is upper bounded by
\begin{align}
\frac{1}{\eta \norm{\mu}} \frac{k\alpha - \la u ,z \ra}{\la \mhat , z\ra } = \frac{k\alpha - \la u ,z \ra}{\eta \la \mu , z\ra } \leq \frac{\alpha}{\eta \la \mu , z \ra}.
\end{align}
We know that if $\la z_i, z_j \ra \geq 0$ for all $z_i,z_j \in Z$ then we only update from $t_{k-1}$ to $t_k$ after we have projected $u$ on the hyperplane $H_{z}^{k\alpha}$ in the direction $\mu$ for all $z \in S_{t_{k-1}}$.  Therefore in this case, the total time taken before $\beta(t_{k-1})$ updates to $\beta(t_k)$ is upper bounded by
\begin{equation}
t_k - t_{k-1} \leq \left( \sup_{m\geq 0} |S_m| \right) (t_s - t_{s-1}) \leq   \left( \sup_{m \geq 0} |S_m| \right) \sup_{t\geq 0} \min_{z \in Z} \abs{\frac{ \alpha }{\eta \sum_{z' \in S_t} \la z,z' \ra}}   .
\end{equation}
and since $S_{k_t} \subset Z$, we obtain
\begin{equation}\label{ktoriginal}
t_k - t_{k-1}  \leq   \sup_{t\geq 0} \min_{z \in Z} \abs{\frac{n \alpha }{\eta \sum_{z' \in S_t} \la z,z' \ra}} 
\end{equation}
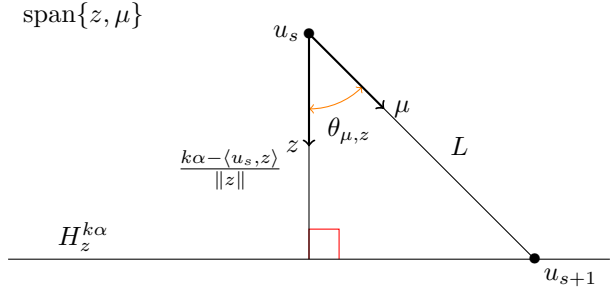
\begin{figure}
\center
\begin{tikzpicture}
    \draw[red] (0,-2) -- (0,-1.6) -- (0.4,-1.6) -- (0.4,-2) ;
  \draw
    (3,-2) coordinate (m) node[below right] {$u_{s+1}$}
    -- (0,1) coordinate (u0) node[left] {$u_s$}
    -- (0,-2) coordinate (z) node[above right] {}
    pic["$\theta_{\mu,z}$", draw=orange, <->, angle eccentricity=1.4, angle radius=1cm]
    {angle=z--u0--m};
    \draw
    (-4,-2) -- (4,-2);
    \draw[thick,->] (0,1) -- (0,-0.5) node[left] {$z$};
    \draw[thick,->] (0,1) -- (1,0) node[right] {$\mu$};
    \draw (2,-0.5) node {$L$};
    \draw (-1.05,-0.8) node {$\frac{k\alpha - \la u_s, z \ra }{\norm{z}}$};
    \draw (-3,-1.7) node {$H^{k\alpha}_{z}$};
    \draw (-3,1.25) node {$\spa \{z,\mu\}$};
    \node at (3,-2) {\textbullet};
    \node at (0,1) {\textbullet};
\end{tikzpicture}
\caption{The right triangle that appears in $\spa \{z,\mu\}$ as a result of projection onto the hyperplane $H^{k\alpha}_{z}$ in the direction $\mu$. The distance we are interested in is the length of the hypotenuse $L$.}
\label{fig:geometry}
\end{figure}
However, if there exists $z_i,z_j \in Z$ such that $\la z_i, z_j \ra < 0$ then it follows there may exist gradients such that $\langle \nabla \hcR(u_{s-1}), \nabla \hcR(u_s) \rangle < 0$.  In this case it follows $-\nabla \hcR(u_{s-1}) = -\nabla \hcR(u_s) + z_i$ for some $z_i \in Z$, but $\langle -\nabla \hcR(u_{s}), z_i \rangle < 0$ .  This may occur when we move in the direction $-\nabla \hcR(u_s)$ towards a hyperplane $H^{k\alpha}_{z}$ which $u_t$ has already crossed, causing us to bounce back and forth over the hyperplane $H^{k\alpha}_{z_i}$ until $\la u_k,z_i \ra > k\alpha$ is satisfied along with the conflicting constraints.  This is because the inner products of the gradients on either side of the hyperplane $H^{k\alpha}_{z}$ is negative causing a clash.  This implicitly causes the gradient flow to move in the direction of the projection of $-\nabla \hcR(u_s)$ onto $H^{k\alpha}_{z_i}$. Let $v = \pi_{z_i} (- \nabla \hcR(u_s) )$, then the length of this line segment in $H^{k\alpha}_{z}$ is given by  $d_{z} = \min_{1 \leq r \leq n} \frac{k \alpha - \la u, z_r \ra }{ \la v, z_r \ra } \leq \min_{1 \leq r \leq d} \frac{\alpha}{\la v, z_r \ra } $.  We know $ \max_r \la v, z_r \ra > 0$ because if not, then $v$ never crosses any hyperplanes $H^{k\alpha}_{z_r}$ and we move in this direction indefinitely, which contradicts linear separability and convexity of the hinge loss.  So we can set $K_{Z} = \max_{1 \leq r \leq n}\la v, z_r \ra $.   Then the distance is bounded by $ \alpha /K_Z$, and the number of gradient steps we take to traverse this length is upper bounded by 
\begin{equation} 
\frac{\alpha /K_Z}{\eta  \frac{1}{ \norm{v}} \min \left\{  \langle - \nabla \hcR(u_s)  , v\rangle , \langle  -\nabla \hcR(u_s) +  z_i , v\rangle  \right\} } =  \frac{\alpha  }{\eta \langle -\nabla \hcR(u_s), \hat v \rangle   K_Z}  .
\end{equation} 
This is because we move in the direction $-\nabla \hcR(u_s) $ part of the time and  $- \nabla \hcR(u_s) + z_i$ the rest of the time.  So simply taking the minimum of the two provides an upper bound.  However, we also have that the two quantities are equal under inner product with $v$, since $\la v,\gamma_i \ra = 0$. We have
\begin{align*}
\la -\nabla \hcR(u_s),  v \ra  = \norm{\nabla \hcR(u_s)}^2 - \frac{1}{\norm{z_i}^2} \la  \nabla \hcR(u_s), z_i \ra^2 
\end{align*}
and so $\la- \nabla \hcR(u_s), \hat v \ra $ is lower bounded by a constant $L_Z > 0$, and 
\begin{equation} \label{dualtime}
\frac{\alpha  }{\eta \langle -\nabla \hcR(u_s), \hat v \rangle   K_Z} \geq \frac{\alpha }{\eta L_Z K_Z}  .
\end{equation} 

So to obtain a bound for either case we simply take the maximum of the two bounds yielding
\begin{align}
t_k - t_{k-1} \leq \frac{n\alpha}{\eta} \max \left\{ \sup_{t\geq 0} \min_{z \in Z} \abs{\frac{n \alpha }{\eta \sum_{z' \in S_t} \la z,z' \ra}}, \frac{1 }{ L_Z K_Z} \right\}
\end{align}
If $\sum_{z' \in S_t} \la z, z' \ra = 0$ then it follows that the hyperplane $H^{k\alpha}_{z}$ is parallel to $-\nabla \hcR(u)$, and so this case is impossible, and so there must exist a constant $C_Z$ such that
\begin{align}
\inf_{t\geq 0} \max_{z \in Z}  \sum_{z' \in S_t} \la z, z' \ra  \geq C_{Z} > 0.
\end{align} 
and 
\begin{align}
t_k - t_{k-1} \leq  \frac{n\alpha}{\eta}\max \left \{ \frac{1}{ C_{Z}}, \frac{1 }{ L_Z K_Z}  \right\} = \bigO(n).
\end{align}
Note that in the case that $\Gamma = I$ this yields $C_{Z} = \norm{e_i}^2 = 1$.

Thus, as a result the asymptotics is unaffected by this change of time variable, as we only require scaling the time by a constant.  So we can consider $u_{t_k}$ to be our iterates given that we multiply our final convergence rate by $n$.
\end{proof}
\begin{proof}[Proof of \autoref{posgrad}]
Take $z \in Z \setminus S$ such that $(t-1) \alpha \leq \la u ,z \ra \leq k\alpha$, then
\begin{align*}
\la -\pi_{\bar u}(u_t), z \ra &= -\la (I - \bar u \bar u^\top) u_t , z \ra \\
&=  \la \bar u , u_t \ra \la \bar u , z \ra - \la u_t ,z \ra  \\
&\geq (\gamma  + \epsilon )\la \bar u , u_t \ra - \la u , z \ra \\
&\geq (\gamma + \epsilon )\la \bar u, u_t \ra - k\alpha \\
&= \gamma(\gamma + \epsilon)  \sum_{i=1}^d \la u_t , \gamma_i^* \ra - k\alpha \\
&= \gamma(\gamma + \epsilon) \sum_{i=1}^d \la \sum_{j=1}^d \la u_t, \gamma_j \ra \gamma_j^*, \gamma_i^* \ra  - k\alpha \\
&= \gamma(\gamma + \epsilon) \sum_{i=1}^d \sum_{j=1}^d \la  \la u_t, \gamma_j \ra \gamma_j^*, \gamma^*_i \ra  - k\alpha \\
&\geq \gamma (\gamma + \epsilon) (k-1)\alpha \sum_{i=1}^d  \sum_{j=1}^d \la \gamma_j^*, \gamma^*_i \ra   - k\alpha ~~ \fbox{By Lemma \ref{dualpos}} \\
&=  \gamma (\gamma + \epsilon) \norm{ \sum_{i=1}^d \gamma_i^*}^2 (k-1)\alpha  - k\alpha \\
&= \frac{\gamma + \epsilon}{\gamma } (k-1)\alpha  - k\alpha .
\end{align*}
From this it follows that if $t > \frac{\gamma + \epsilon}{\epsilon} +K $ and $(k-1)\alpha \leq \la u ,z \ra \leq k\alpha$ then
\begin{align*}
\la -\pi_{\bar u}(u), z \ra \geq K\alpha.
\end{align*}
\end{proof}
In order to construct the polytope $R_{d}(k)$ we must first define some preliminary objects which will be used in the construction.  First we construct an operator $\phi : \mathbb{R}^d \to \mathbb{R}^d$ such that
\begin{align*}
\phi(u) = \min_{z \in Z} \frac{ \lceil  \la u ,z \ra / \alpha \rceil \alpha - \la u, z \ra }{\la- \nabla \hcR(u), z \ra } 
\end{align*}
Geometrically speaking the operator $\phi$ is the projection of $u$ onto the hyperplane $H^{k\alpha}_{z}$ in the direction $-\nabla \hcR(u)$, and it has the property that $u_s = u_{s-1} +\phi(u_{s-1}) $.  Given $\phi$ we can construct another operator $\psi$ as follows
\begin{align*}
\psi(u) = \arg \min_{\gamma_i \in \Gamma} \{ t ~|~  \phi^{t}(u) = \gamma_i \}
\end{align*}
Thus, the operator $\psi$ gives us the first $\gamma_i$ whose corresponding hyperplane $H^{k\alpha}_{\gamma_i}$ we cross during gradient descent.
\begin{definition}[Polytope Construction]\label{construction}
First we choose a support matrix $\Gamma$ such that for all $\gamma_i \in \Gamma$ we have that for all $z \in S \setminus \Gamma$ that $z$ can be written as a convex combination of $\gamma_i$.  Because $\sum_{i=1}^d \gamma_i^* = \bar u /\gamma$ for any support matrix, we can do this by simply maximizing $\abs{\det(\Gamma)}$. This quantity gives the volume of the parallelotope generated by $(\gamma_i)_{i=1}^m$.  So we choose a support matrix $\Gamma$ such that
\begin{align*}
\Gamma = \arg \max_{\Gamma} \abs{\det (\Gamma) }.
\end{align*}
Then given any submatrix $ (\gamma_1,\dots,\gamma_m)^\top\in \mathbb{R}^{m \times d}$ of $\Gamma$ we define $E_{m}(v,k,i)$ with $v \in \{0,1\}^m$, $1 \leq m \leq d$, $ k \geq 0$, and $i \in \{1,\dots,m\}$ as the set 
\begin{align*}
E_{m}(v,k,i) = \left\{ u \in \mathbb{R}^d ~|~ \psi(u) = \gamma_i,~ \alpha((k-1)\ones +v) \preceq \Gamma u \preceq \alpha(k\ones+v) \right\}
\end{align*}
Then we can define
\begin{align*}
R_m(k) = \bigcup_{\sigma \in \mathfrak{S}_m} \bigcup_{r=1}^d E_{m}\left( \sum_{i=1}^r e_{\sigma(i)}, k, r+1\right).
\end{align*}
Where $\mathfrak{S}_m$ is the set of all permutations of $m$ objects.
\end{definition}
Restricting ourselves to the smallest $\Gamma$ allows us to ensure that no other support matrix is contained within $\Gamma$, and therefore the optimization problem within the parallelotope generated by $(\gamma_i)_{i=1}^m$ is the same as a vanilla hinge loss problem where $P(k+1)$  contains all critical points.  See Figure \ref{fig:reducetosingle} for a visualization.  \autoref{construction} also gives us access to a family of polytopes $R_m(k)$ for $m \leq d$ which we will also use in the following proof of \autoref{polytope}.
\begin{proof}[Proof of \autoref{polytope}]
By \autoref{posgrad} we can assume $S_k \subset S$ for all $k$.

First we prove property 1.  We can prove this even with the tightest possible superset.  Let $\bar u$ be the max-margin separator, then 
we have for $1 \leq i \leq d$ that
\begin{align*}
\la \frac{k\alpha}{\gamma} \bar u,\gamma_i \ra  = k\alpha \la \sum_{j=1}^d \gamma_j^*, \gamma_i \ra = k \alpha
\end{align*}
So then we know that $\frac{k\alpha}{\gamma} \bar u \in P(k) \subset R_d(k)$.

For property 2, we apply the construction of $R_{m}(k)$ from \autoref{construction}.  Let $u_k \in R_m(k)$, then we can assume without loss of generality that $u_k \in \bigcup_{r=0}^d E_m \left( \sum_{i=1}^r e_i , k , r+1 \right)$  (i.e. $\sigma$ is the identity permutation $e$) because if not then we can simply permute the indexing of the $\gamma_i$ such that it is true.  Suppose $z \notin \Gamma$ corresponds with the next hyperplane $H^{k\alpha}_{z}$ crossed during gradient descent, then
\begin{align*}
u_s = u_{s-1} + \frac{k \alpha - \la u_{s-1}, z \ra }{\la \nabla \hcR(u_{s-1}), z \ra } \nabla \hcR(u_{s-1})
\end{align*}
So for $i > r$,
\begin{align*}
\la u_s, \gamma_i \ra &= \la u_{s-1}, \gamma_i \ra + \frac{k \alpha - \la u_{s-1}, z \ra }{\la \nabla \hcR(u_{s-1}), z \ra } \la  \nabla \hcR(u_{s-1}), \gamma_i \ra \\
&\leq  \la u_{s-1}, \gamma_i \ra + \frac{k \alpha - \la u_{s-1}, \gamma_i \ra }{\la \nabla \hcR(u_{s-1}), \gamma_i \ra } \la  \nabla \hcR(u_{s-1}), \gamma_i \ra \\
&= k\alpha ,
\end{align*}
and for $i \leq r$
\begin{align*}
\la u_s, \gamma_i \ra &= \la u_{s-1},\gamma_i \ra  + \frac{k \alpha - \la u_{s-1}, z \ra }{\la \nabla \hcR(u_{s-1}), z \ra } \la  \nabla \hcR(u_{s-1}), \gamma_i \ra  \\
&\leq \la u_{s-1},\gamma_i \ra + \frac{(k+1) \alpha - \la u_{s-1}, \gamma_i \ra }{\la \nabla \hcR(u_{s-1}), \gamma_i \ra }  \la \nabla \hcR(u_{s-1}), \gamma_i \ra \\
&= (k+1)\alpha.
\end{align*}
If on the other hand $z \in \Gamma$, then if $z = \gamma_i$ with $i > r$ then it follows $i  = r+1$ by construction and
\begin{align*}
\la u_s,\gamma_i \ra  &= \la  u_{s-1},\gamma_i \ra + \frac{k\alpha - \la u_{s-1},\gamma_i \ra }{\la \nabla \hcR(u_{s-1}), \gamma_i \ra } \la \nabla \hcR(u_{s-1}) , \gamma_i \ra  \\
&= k \alpha.
\end{align*}
In this case we have ``upgraded'' from $E_m(\sum_{i=1}^{r} e_i,k,r+1)$ to $E_m(\sum_{i=1}^{r+1} e_i,k,r+2)$.  On the other hand if $z = \gamma_i$ with $i \leq r$ then we go backwards and
\begin{align*}
\la u_s,\gamma_i \ra  &= \la  u_{s-1},\gamma_i \ra + \frac{k\alpha - \la u_{s-1},\gamma_i \ra }{\la \nabla \hcR(u_{s-1}), \gamma_i \ra } \la \nabla \hcR(u_{s-1}) , \gamma_i \ra  \\
&= k \alpha.
\end{align*}
In this case, we end up ``downgrading'' to $E_m( \sum_{j=1}^r e_j - e_i, k, r+1 ) $.  However, we also have $-\nabla \hcR(u_s) = -\nabla \hcR(u_{s-1}) + \gamma_i$, and so eventually we will bounce back and forth along the hyperplane $H^{k\alpha}_{\gamma_i}$ until the constraint $\la u , \gamma_i \ra \geq k\alpha$ is satisfied along with the future constraints $\la u, \gamma_j \ra $ where $ j > r$. The amount of additional time this takes is bounded as we have seen in eq. \ref{dualtime}.

So then it follows that by induction, eventually we will reach $u_{s} := u_{k+1}$ with $ k \alpha \leq \la u_{k+1}, z \ra \leq (k+1)\alpha$ and hence, $u_s = u_{k+1} \in R_{d}(k+1)$, as eventually we will have $\la u_s, \gamma_i \ra \geq k\alpha$ for all $i \leq m$ and $\la u_s, \gamma_i \ra \leq (k+1)\alpha$.

Now we prove property 3.
By Lemma \ref{posgrad} and since $\bar u^\perp \subset \bigcup_{k \geq 0} R_{d}(k) $ we know that the addition of non-support vectors $z \in Z$ will only accelerate convergence to $R_{d}(k)$, so we can assume that we only train on support vectors for this lemma. We know that $u_0$ is always in $R_m(k)$ for some $m\leq d$ because there will always be at least one $\gamma_i \in \Gamma$ with $\la u,\gamma_i \ra \leq k\alpha$ for some support matrix $\Gamma$ (not necessarily the minimal one from \autoref{construction}).  Therefore, we prove this by induction on $m$.  Note that in the one dimensional case we obtain
\begin{align*}
R_1(k) =  \bigcup_{v \in \{0,1\}} E_m(v,k,1) = E_m (0,k,1) \cup E_m (1,k,1).
\end{align*}
Which implies our polytope is a single line segment.  So then $R_1(k)$ is simply a ray starting at $u_0$ traveling in a single direction $\gamma_1 \in \Gamma$.

Now we argue that if $u_0 \in R_{m-1}(k)$ then after a constant time $T_m$ we have that $u_{T_m} \in R_{m}(k+T_m)$.  Let $\Gamma_m \in \mathbb{R}^{m \times d}$ be the first $m$ rows of $\Gamma$ and let $\bar u_m = \frac{1}{\norm{\Gamma_m^\dagger \ones_m}} \Gamma_m^\dagger \ones$. Then if $u_0 \in R_{m-1}(k)$ then the distance between $u_{k}$ and $ k \alpha \norm{\Gamma^\dagger_{m-1} \ones_{m-1}} \bar u_{m-1} = k \alpha \Gamma^\dagger_{m-1} \ones_{m-1}$ is bounded by the constant $C_{m-1} = \sup_{x,y \in R_{m-1}(k)} \norm{x-y} $, so we can simply upper bound the distance between $u_0$ and $u_{T_k}$ in terms of the distance $u_0$ travels in the direction $ \bar u_{m-1}$ to reach the hyperplane parallel to $\bar u$.  Then we can calculate that the two rays cross each other some time before traversing the distance
\begin{align*}
D_m &=  \frac{\la  \pi_{\bar u_m}(u_0), u_0 \ra }{\la  \pi_{\bar u_m}(u_0), \bar u_{m-1} \ra }.
\end{align*}
We know that $ \la \pi_{\bar u_m}(u_0) , \bar u_{m-1} \ra \neq 0$ because if not then it follows that $\bar u_{m-1} = \bar u_{m}$, and we must have that $u_0 \in R_m(k)$ by default.  Thus, we can bound $T_m \leq \sup_{s\geq 0} D_m / (\eta \Vert\nabla \hcR(u_s)\Vert)$.   

Now we can simply sum up the constants $T_m$ to obtain that the total time $T \leq \sum_{m=1}^d T_m$ which is still a constant. Thus, by induction we know that if $u_0 \in \bigcup_{k > 0} R_m(k)$ for any $1 \leq m \leq d$ then eventually for some $T$, $u_{T} \in \bigcup_{k>0} R_d(k)$.  
\end{proof}
\begin{proof}[Proof of \autoref{stayahead2}]
We know that $\frac{k\alpha}{\gamma} \bar u \in R_{d}(k)$, and that $C = \sup_{x,y \in R_{d}(k)} \norm{x-y}$ is bounded. So then it follows if $u \in R_{d}(k)$ then 
\begin{align*}
u = \frac{k\alpha}{\gamma} \bar u +  v
\end{align*}
for some $v \in \mathbb{R}^d$ with $\norm{v} \leq C$. Then it follows
\begin{align*}
\la  u, z\ra &= \la \frac{k\alpha}{\gamma} \bar u + v ,z \ra \\
&\geq \frac{\gamma+\epsilon}{\gamma} k\alpha +  \la v, z \ra \\
&\geq \frac{\gamma + \epsilon}{\gamma} k\alpha - \norm{v}\norm{z} \\
&\geq \frac{\gamma + \epsilon}{\gamma} k\alpha - C \norm{z}.
\end{align*}
So then it follows for $k > \frac{\gamma C \norm{z}}{\alpha \epsilon}$, that $\la u ,z \ra \geq k\alpha$.
\end{proof}
\begin{proof}[Proof of \autoref{linearparamconv}]
Apply Lemmas \ref{posgrad}, \ref{polytope}, and \ref{stayahead2}, then it follows that for $k> \frac{\gamma + \epsilon}{\epsilon}+T+ \frac{\gamma C \norm{Z}_{1}}{\alpha \epsilon}$,
\begin{align}
\norm{u_k - \frac{k\alpha}{\gamma} \bar u } \leq \sup_{x,y \in R_d(k)} \norm{x-y}.
\end{align}
Because $R_d(k)$ is bounded with $C = \sup_{x,y \in R_{d}(k)} \norm{x-y}$ then, it follows
\begin{align*}
\norm{u_k}  &= \norm{u_k - \frac{k\alpha}{\gamma}\bar u + \frac{k\alpha}{\gamma} \bar u} \\
&\leq \norm{u_k - \frac{k\alpha}{\gamma} \bar u } + \norm{\frac{k\alpha}{\gamma} \bar u} \\
&\leq C + \frac{k\alpha}{\gamma}.
\end{align*}
We know that $\min_{z \in Z} \la u_k, z \ra \geq k\alpha$ by definition of the complete hinge loss.  Then we have
\begin{align*}
\gamma  - \min_{ z\in Z} \la \frac{u_k}{\norm{u_k}} , z \ra  &\leq \gamma -  \frac{(k-1)\alpha }{C + \frac{k\alpha}{\gamma}} \\
&= \bigO \left( \frac{1}{k} \right)
\end{align*}  
Then we apply Lemma \ref{subseq} to switch back to the original gradient descent iterates to obtain
\begin{align*}
\gamma  - \min_{ z\in Z} \la \frac{u_t}{\norm{u_t}} , z \ra = \bigO \left( \frac{n}{t} \right).
\end{align*}
Additionally,
\begin{align*}
1 - \la \frac{u_k}{\norm{u_k}}, \bar u \ra &= 1 - \frac{1}{\norm{u_k}} \la \sum_{i=1}^d \la u_k,\gamma_i \ra  \gamma_i^* , \bar u \ra \\
&\leq  1 - \frac{(k-1)\alpha}{\norm{u_k}} \la  \sum_{i=1}^d \gamma_i^* , \bar u \ra \\
&=1 - \frac{ ((k-1)\alpha) /\gamma}{\norm{u_k}} \\
&\leq  1 - \frac{ ((k-1)\alpha) /\gamma}{C + \frac{k\alpha}{\gamma}} \\
&= \bigO \left( \frac{1}{k} \right),
\end{align*}
and so
\begin{align*}
1 - \la \frac{u_t}{\norm{u_t}}, \bar u \ra &= \bigO \left( \frac{n}{t} \right).
\end{align*}
And lastly,
\begin{align*}
\norm{ \frac{u_k}{\norm{u_k}} - \bar u}^2 &= 2 - 2 \frac{1}{\norm{u_k}}\la  \bar u, u_k \ra  \\
&\leq 2 - 2 \frac{(k-1)\alpha/\gamma}{C + \frac{k\alpha}{\gamma}} \\
&= \bigO \left( \frac{1}{k} \right),
\end{align*}
and so
\begin{align*}
\norm{ \frac{u_t}{\norm{u_t}} - \bar u} = \bigO \left( \sqrt{\frac{n}{t}} \right).
\end{align*}
\end{proof}
\section{Extra Experiment Details}\label{appendixB}
\begin{figure}[H]
\centering
\begin{tabular}{lccccc}
\textbf{Layer} & \textbf{Filters} & \textbf{Size} & \textbf{Stride} & \textbf{Activation} & \textbf{Times}\\
\hline
Conv1 &  64 & $3 \times 3$ & $1 \times 1$ & ReLU & 1\\
\hline
Conv2 &  128 & $3 \times 3$ & $1 \times 1$ & ReLU & 3\\
\hline
MaxPool1 &  - & $2 \times 2$ & $2 \times 2$ & - & 1\\
\hline
Conv3 &  128 & $3 \times 3$ & $1 \times 1$ & ReLU & 3 \\
\hline
MaxPool2 &  - & $2 \times 2$ & $2 \times 2$ & - & 1 \\
\hline
Conv4 &  128 & $3 \times 3$ & $1 \times 1$ & ReLU & 2 \\
\hline
MaxPool3 &  - & $2 \times 2$ & $2 \times 2$ & - & 1 \\
\hline
Conv5 &  128 & $3 \times 3$ & $1 \times 1$ & ReLU & 1 \\
\hline
Conv6 &  128 & $1 \times 1$ & $1 \times 1$ & ReLU & 2 \\
\hline
MaxPool4 &  - & $2 \times 2$ & $2 \times 2$ & - & 1 \\
\hline
Conv7 &  128 & $1 \times 1$ & $1 \times 1$ & - & 1
\end{tabular}
\caption{Architecture specification for convolutional neural network used on CIFAR-10 in \autoref{neural}.  ``Times'' column refers to the number of times the layer was repeated.  All convolutional layers were directly followed by a batch norm layer.}
\end{figure}
For the experiments with MNIST and CIFAR-10 we extend the binary risk function $\hcR(f,\beta)$ to a multiclass risk function $\hcR_{\textup{multi}} (f,\beta)$ using the strategy proposed by \cite{weston1999}. That is we define $\hcR_{\textup{multi}}(f,\beta)$ as
\begin{align*}
\hcR_{\textup{multi}} (f,\beta) &:= \frac{1}{n}\sum_{i=1}^n   \sum_{y \neq y_i } \1[ f(x_i)_{y_i} - f(x_i)_{y}  \leq \beta](f(x_i)_{y} - f(x_i)_{y_i}) \\  
&- \1 \left[ \sum_{i=1}^n \sum_{y \neq y_i}   \max \{ \beta - (f(x_i)_{y_i} - f(x_i)_{y}) , 0 \} \leq \zeta \right]  \frac{\alpha}{n\eta } \beta  .
\end{align*}
\end{document}